\documentclass[10pt,twocolumn,letterpaper]{article}

\usepackage{iccv}              

\usepackage[accsupp]{axessibility}  

\usepackage{graphicx}
\usepackage{booktabs}
\usepackage{amsmath}
\usepackage{amssymb}
\usepackage{xcolor}
\usepackage{color}
\usepackage{bbding}
\usepackage{array,multirow,textcomp}
\usepackage{rotating}
\usepackage[skip=0.5\baselineskip]{caption}
\usepackage{makecell}
\usepackage{booktabs}
\usepackage{adjustbox}
\usepackage{array}
\usepackage{balance}
\usepackage{wrapfig}
\usepackage{enumitem}
\usepackage{tikz}
\usepackage{tabularx}
\usepackage{colortbl}
\usepackage{etoolbox}
\usepackage{placeins}
\usepackage{float}
\usepackage{pgf}
\usepackage{subcaption}
\usepackage{arydshln}
\usepackage{fontawesome}
\usepackage[dvipsnames]{xcolor}

\usepackage{tikz}

\newcolumntype{Y}{>{\centering\arraybackslash}X}
\newcolumntype{C}[1]{>{\centering\let\newline\\\arraybackslash\hspace{0pt}}m{#1}}
\newcolumntype{R}[2]{%
    >{\adjustbox{angle=#1,lap=\width-(#2)}\bgroup}%
    l%
    <{\egroup}%
}

\newcommand*\rots{\multicolumn{1}{R{60}{1em}}}

\definecolor{iccvblue}{rgb}{0.21,0.49,0.74}
\usepackage[pagebackref,breaklinks,colorlinks,allcolors=iccvblue]{hyperref}

\makeatletter
\renewcommand*{\@fnsymbol}[1]{\ensuremath{\ifcase#1\or *\or \dagger\or \ddagger\or
    \mathsection\or \mathparagraph\or \|\or **\or \dagger\dagger
    \or \ddagger\ddagger \else\@ctrerr\fi}}
\makeatother

\title{Unlocking Constraints: Source-Free Occlusion-Aware Seamless Segmentation}

\author{
Yihong Cao$^{1,2,*}$ \quad Jiaming Zhang$^{3,4,*}$ \quad Xu Zheng$^{5,6}$ \quad Hao Shi$^{7}$ \quad Kunyu Peng$^{3}$ \quad Hang Liu$^{1}$\\Kailun Yang$^{1,\dagger}$ \quad Hui Zhang$^{1,\dagger}$\\
$^{1}$Hunan University \quad $^{2}$Hunan Normal University \quad $^{3}$Karlsruhe Institute of Technology \quad $^{4}$ETH Z\"urich\\$^{5}$HKUST(GZ) \quad $^{6}$INSAIT, Sofia University ``St. Kliment Ohridski'' \quad $^{7}$Zhejiang University
}

\begin{document}

\twocolumn[{
\renewcommand\twocolumn[1][]{#1}%
\maketitle
\begin{center}
    \centering
    \captionsetup{type=figure}
    \includegraphics[width=\textwidth]{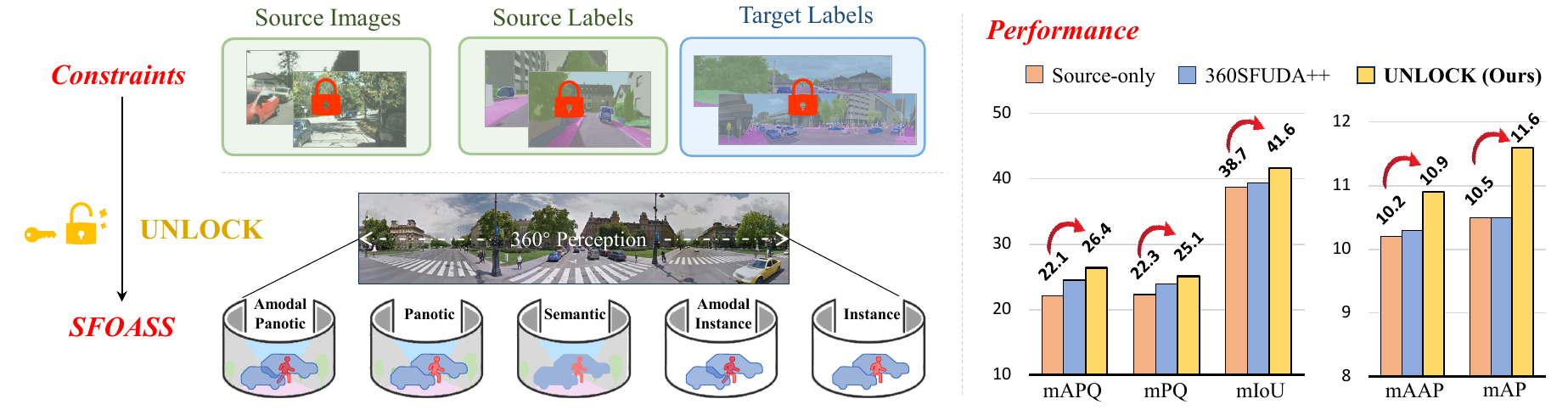}
    
    \begin{minipage}[t]{.6\textwidth}
        \vskip -2.5ex
        \subcaption{
        Unlocking the constraints of SFOASS with the proposed UNLOCK framework.
        }\label{fig1a}
    \end{minipage}%
    \begin{minipage}[t]{.4\textwidth}
        \vskip -2.5ex
        \subcaption{Performance comparison of SFOASS.}\label{fig1b}
    \end{minipage}%
    \setcounter{figure}{0}

    \captionof{figure}{\textbf{UNLOCK} framework \textbf{\textit{(a)}} solves the Source-Free Occlusion-Aware Seamless Segmentation (SFOASS), enabling segmentation with 360{\textdegree} viewpoint coverage and occlusion-aware reasoning while adapting without requiring source data and target labels, and \textbf{\textit{(b)}} outperforms existing SFDA methods on the Real-to-Real scenario~\cite{cao2024oass} across all five segmentation metrics (\ie, mAPQ for amodal panoptic, mPQ for panoptic, mIoU for semantic, mAAP for amodal instance, and mAP for instance).}
    \label{fig1:banner}
\end{center}%
}]

{  \renewcommand{\thefootnote}{$^\ast$}
  \footnotetext{Equal contribution. $^\dagger$Corresponding authors (e-mail: {\tt kailun.yang@hnu.edu.cn, zhanghuihby@126.com}).}
}

\begin{abstract}
\vskip -3.1ex
Panoramic image processing is essential for omni-context perception, yet faces constraints like distortions, perspective occlusions, and limited annotations. Previous unsupervised domain adaptation methods transfer knowledge from labeled pinhole data to unlabeled panoramic images, but they require access to source pinhole data. To address these, we introduce a more practical task, \ie, Source-Free Occlusion-Aware Seamless Segmentation (SFOASS), and propose its first solution, called \textbf{UNconstrained Learning Omni-Context Knowledge (UNLOCK)}. Specifically, UNLOCK includes two key modules: Omni Pseudo-Labeling Learning and Amodal-Driven Context Learning. While adapting without relying on source data or target labels, this framework enhances models to achieve segmentation with 360{\textdegree} viewpoint coverage and occlusion-aware reasoning. Furthermore, we benchmark the proposed SFOASS task through both real-to-real and synthetic-to-real adaptation settings. Experimental results show that our source-free method achieves performance comparable to source-dependent methods, yielding state-of-the-art scores of 10.9 in mAAP and 11.6 in mAP, along with an absolute improvement of +4.3 in mAPQ over the source-only method. All data and code will be made publicly available at \url{https://github.com/yihong-97/UNLOCK}.
\end{abstract}    
\section{Introduction}
Scene understanding is a foundational task in computer vision, essential for various downstream applications like autonomous driving~\cite{ma2021densepass, yang2021capturing}, virtual reality~\cite{serrano2019motion, davidson2020360}, and robotics~\cite{bacchin2024pannote,shi2023panoflow}. 
Despite significant advances in the field, current methods still face challenges in achieving both comprehensive and human-like perceptions of surrounding environments, limiting their ability to fully understand scenes. Multiple critical research questions remain unsolved. 

\noindent \textbf{\textit{1) How to efficiently expand to 360{\textdegree} field of view?}} 
Traditional scene understanding methods~\cite{chen2018deeplabv3+, xie2021segformer, guo2022segnext} tailored for pinhole imagery, are limited by a narrow Field of View (FoV).
In contrast, panoramic vision offers 360{\textdegree} view in a single shot, which has the potential to facilitate more comprehensive scene understanding~\cite{gao2022review,ai2022deep_omnidirectional}. 
However, directly applying traditional methods to panoramic data often results in unreliable performance due to challenges such as distortions~\cite{zheng2024ops}.
For effective panoramic understanding, it is essential to learn omni-context knowledge when transferring from the pinhole domain to the panoramic domain. 
This knowledge transfer is crucial for accurate feature interpretation and wide-FoV representation, ultimately enabling seamless 360{\textdegree} scene understanding~\cite{zhang2022bending,zhang2021transfer}.

\noindent \textbf{\textit{2) How can amodal prediction be improved to extend the depth of view?}} 
Amodal perception~\cite{li2016amodal_instance_segmentation, qi2019amodal_kins}, the ability to predict an occluded object's complete shape, is key to robust scene understanding~\cite{zhang2021issafe} and mirrors human-like perception. 
Extending the depth of view (vertical to field of view) means that models can achieve seamless perception, which refers to pixel-level segmentation of the background and occlusion-aware instance-level segmentation of the foreground. 
Beyond the previous amodal segmentation methods~\cite{zhu2017semantic_amodal_segmentation,follmann2019learning_see_invisible}, how to improve occlusion-aware reasoning like OASS~\cite{cao2024oass} in both directions (depth and field of view) simultaneously remains challenging.

\noindent \textbf{\textit{3) How can models adapt to panoramic domains without source data?}} Apart from the field and depth of view, the scarcity of labeled data presents a significant limitation in panoramic vision.
To address the lack of labeled data, Unsupervised Domain Adaptation (UDA) leverages labeled data from the source domain while training on unlabeled data from the target domain~\cite{tranheden2021dacs,jang2022dada_uda,hoyer2022hrda,hoyer2022daformer,jiang2023domain,cao2024oass}.
However, UDA methods still rely on access to source domain, which can often be challenging or even impossible due to privacy concerns or commercial restrictions on source datasets, such as in autonomous driving~\cite{liu2021source,zheng2024semantics_style_matter,zheng2024360sfuda++}.
This restriction leads us to source-free learning, \ie how to adapt models to panoramic domains without source data. 
However, it remains under-explored in amodal panoramic vision.

In this work, to address the aforementioned challenges in scene understanding, we extend the OASS~\cite{cao2024oass} task and introduce a more practical task, \textit{i.e.}, Source-Free Occlusion-Aware Seamless Segmentation (SFOASS). To tackle this task, we propose its first solution, called \textbf{UNconstrained Learning Omni-Context Knowledge (UNLOCK)}. As shown in Fig.~\ref{fig1a}, UNLOCK improves panoramic seamless segmentation performance under adaptation constraints without requiring source data and target labels.
Specifically, to effectively capture the domain-invariant knowledge from the source domain, a method called Omni Pseudo-Labeling Learning (OPLL) is proposed.
Besides, to learn intra-domain knowledge and integrate the domain-invariant knowledge, we put forward the Amodal-Driven Context Learning (ADCL) strategy.
Together, OPLL and ADCL adapt the source model trained on pinhole data to target panoramic images while enhancing panoramic perception and occlusion-aware reasoning.
With OPLL as the key and ADCL performing the unlocking action, they work in synergy to unlock the constraints in SFOASS, thereby enabling seamless perception of the model towards panoramic data.

Aside from Real-to-Real source-free adaptation on the \textit{KITTI360-APS}${\rightarrow}$\textit{BlendPASS} benchmark~\cite{cao2024oass}, we pioneer Synthetic-to-Real adaptation in OASS and SFOASS and introduce \textit{AmodalSynthDrive}${\rightarrow}$\textit{BlendPASS} to investigate the potential of adapting synthetic panoramic images to the real panoramic image domain.
Extensive experiments are conducted on both SFOASS benchmarks. Without access to the original source data (neither images nor labels), our UNLOCK framework can obtain on-par performance as compared to UDA-based methods that require all source data.  
Surprisingly, as shown in Fig.~\ref{fig1b}, results on the Real-to-Real scenario~\cite{cao2024oass} show an absolute improvement of $+4.3$ in mAPQ, reaching $26.4$, compared to the source-only method, and outperforms UDA methods using $12K$ images in instance-level segmentation, with state-of-the-art scores of $10.9$ in mAAP and $11.6$ in mAP, respectively.

In this work, we propose contributions as follows:
\begin{itemize}
    \item We introduce the Source-Free Occlusion-Aware Seamless Segmentation (SFOASS) task and its first solution, UNconstrained Learning Omni-Context Knowledge (UNLOCK). The UNLOCK framework enables models to achieve segmentation with 360{\textdegree} viewpoint coverage and occlusion-aware reasoning while adapting without the need for source data and target labels.
    \item To fully exploit the knowledge from source and target domains, Omni Pseudo-Labeling Learning (OPLL) and Amodal-Driven Context Learning (ADCL) strategies are introduced in UNLOCK to achieve occlusion-aware seamless segmentation for panoramic images. 
    \item Extensive experiments conducted on two SFOASS scenarios, Real-to-Real and Synthetic-to-Real, demonstrate the effectiveness of the UNLOCK framework, highlighting its ability to achieve competitive performance.  
\end{itemize}
\section{Related Work}
\label{related_work}

\noindent\textbf{Seamless segmentation.}
Seamless segmentation, as introduced by~\cite{cao2024oass}, aims to achieve amodal-level segmentation for large-FoV images, enabling unified, occlusion-aware scene understanding. 
Wide-FoV segmentation on fisheye~\cite{deng2017cnn,shi2023fishdreamer,ye2020universal,sekkat2020omniscape,yogamani2019woodscape} and panoramic images~\cite{guttikonda2023single_spherical,li2023sgat4pass,yang2019pass,yang2021context,zheng2023complementary_bidirectional} facilitates comprehensive 360{\textdegree} scene comprehension~\cite{yang2019can}. 
Panoramic panoptic segmentation enhances scene understanding by providing instance-level insights~\cite{fu2023panopticnerf,jaus2023panoramic_insights,mei2022waymo,porzi2019seamless}.
Amodal segmentation extends this concept by predicting both visible and occluded object regions~\cite{hariharan2014simultaneous,dai2015convolutional,back2022unseen_occlusion,sun2022amodal_bayesian,tran2022aisformer,li2023muva,fan2023rethinking,liu2024blade}. 
Li~\etal~\cite{li2016amodal_instance_segmentation} pioneered amodal instance segmentation, introducing occlusion-aware segmentation through iterative regression. 
Subsequent datasets adapted for amodal segmentation~\cite{zhu2017semantic_amodal_segmentation} and  
occlusion classification~\cite{qi2019amodal_kins} have driven further advances. 
Shape and contour priors~\cite{gao2023coarse,chen2023amodal_expansion,li2023gin,xiao2021amodal_segmentation_prior,li20222d_3d_prior} have further refined segmentation accuracy.
In addition to instance segmentation, amodal panoptic segmentation has been explored~\cite{zhu2017semantic_amodal_segmentation,hu2019sail,breitenstein2022amodal_cityscapes}. 
Mohan~\etal~\cite{mohan2022amodal_panoptic_segmentation,mohan2022perceiving} combined semantic and amodal segmentation for amodal panoptic segmentation.
Recent methods, including pix2gestalt~\cite{ozguroglu2024pix2gestalt}, AISDiff~\cite{tran2024amodal}, and Xu~\etal~\cite{xu2024amodal}, leverage diffusion model priors for amodal segmentation. 
Although seamless segmentation~\cite{cao2024oass} unifies aforementioned segmentation paradigms to form occlusion-aware scene understanding for panoramic vision,
our work emphasizes source-free seamless segmentation, aiming for high accuracy even with restricted access to labeled source data.

\noindent\textbf{Source-free domain adaptation.}
Existing works dealing with domain gaps on semantic segmentation tasks mostly focus on UDA where labeled source domain data is required~\cite{zou2018unsupervised,wang2020differential,zhang2021transfer,zheng2023look_neighbor,zhang2021transfer,zheng2023both_style_distortion,zhang2022bending,Tasar2019ColorMapGANUD,Zhao2023UnsupervisedDA,Guizilini2021GeometricUD,Xu2021NeutralCL}. 
Source-Free Domain Adaptation (SFDA)~\cite{liu2021source,rizzoli2024source,lu2023uncertainty,karim2023c,Wang2023FVPFV,Yi2023WhenSD,Yu2023ACS,Duan2024SourceFreeDA,cao2024towards} is a more practical approach compared with UDA, as it removes the need for access to source data during the adaptation process.
As sub-directions of scene segmentation, domain adaptation of panoptic segmentation is challenging as it must distinguish and integrate semantic and instance information across domains, as highlighted in existing works~\cite{saha2023edaps,martinovic2024mc_panda,huang2021cross,zhang2023unidaformer,mansour2024language_guided}.
To overcome the limitations of the FoV, domain adaptation in panoramic segmentation facilitates comprehensive scene understanding across a full 360{\textdegree} perception, while effectively harnessing the rich knowledge from label-dense pinhole domains.
Researchers have explored both UDA~\cite{zhang2024goodsam,zhang2024goodsam++,jiang2024multi_source_da,zhang2022behind,zhang2021transfer,zheng2023both_style_distortion,zhang2022bending} and SFDA~\cite{zheng2024360sfuda++,zheng2024semantics_style_matter} settings for panoramic segmentation.
Jaus~\etal.~\cite{jaus2023panoramic_insights,jaus2021panoramic_towards} illustrate the need for panoramic panoptic segmentation in autonomous driving.  
Zheng~\etal.~\cite{zheng2024360sfuda++,zheng2024semantics_style_matter} for the first time, explored SFDA in panoramic semantic segmentation.
However, the SFDA challenge remains unexplored for OASS~\cite{cao2024oass}. 
In this work, we introduce the novel SFOASS task and benchmark it across two scenarios. We also proposed the first solution UNLOCK, a novel method designed to unlock constraints posed by SFOASS.

\section{UNLOCK Framework}
\begin{figure*}[ht]
    \centering
    \includegraphics[width=\linewidth]{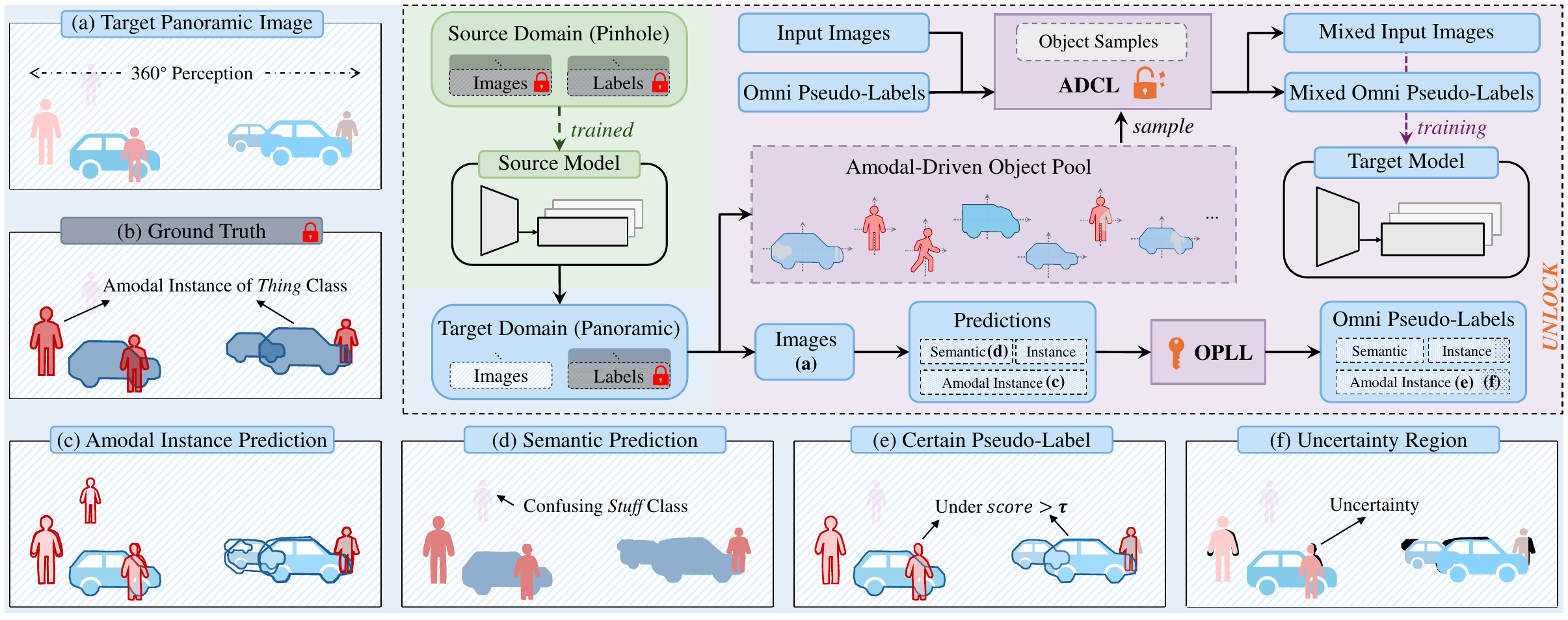}
    \caption{Illustration of the proposed UNLOCK. To address the challenge of the inaccessible source domain and target labels in SFOASS, we propose Omni Pseudo-labeling Learning (OPLL) and Amodal-Driven Contextual Learning (ADCL) modules. OPLL serves as the key, and ADCL, holding the key, unlocks the constraints of SFOASS, enabling effective adaptation of the target model to the panoramic domain.
    }
    \label{fig_arc}
    \vskip-2ex
\end{figure*}
\label{methodology}

\subsection{Overview}
\noindent\textbf{OASS.}
In the OASS~\cite{cao2024oass}, the objective is to use a labeled source pinhole domain $\mathcal{D}^{pin}{=}\{x_{i}^{pin}, y_{i}^{pin}\}_{i=1}^{N^{pin}}$ and an unlabeled target panoramic domain $\mathcal{D}^{pan}{=}\{x_{i}^{pan}\}_{i=1}^{N^{pan}}$ to obtain an OASS model $F$ that performs well in $\mathcal{D}^{pan}$. 
Both domains share the same $C$ categories, which can be further divided into $C^{stu}$ \textit{Stuff} classes and $C^{thi}$ \textit{Thing} classes. 
Ultimately, the $F$ should be able to output five segmentation results with the input panoramic image at once: semantic, instance, amodal instance, panoptic, and amodal panoptic maps. Typically, $F$ is designed to contain a shared encoder and three detection branches: semantic, instance, and amodal instance branches, each outputting corresponding predictions $p_{i}^{sem}, p_{i}^{ins}, p_{i}^{ains}$, denoted as
\begin{equation}
    p_{i}^{sem}, p_{i}^{ins}, p_{i}^{ains} = F(x_{i}),
\end{equation}
where $p_{i}^{sem}{\in}\mathbb{R}^{H\times{W}{\times}C}$ represents the semantic prediction of the input image $x_i{\in}\mathbb{R}^{H{\times}W {\times}3}$. 
$p_{i}^{ins}$ and $ p_{i}^{ains}$ represent the instance and amodal instance predictions, respectively, with each including $class{\in}\mathbb{R}^{C^{thi}}$, $score{\in}\left [0,1\right ]$, and $mask{\in}\mathbb{R}^{H{\times}W}$ predictions for $j$th object.
During adaptation, the label $y_i$ of $x_i$ consists of three components: semantic $y_{i}^{sem}$, instance $y_{i}^{ins}$, and amodal instance $y_{i}^{ains}$ labels, used for supervised learning in their respective branches. In the inference phase, the predictions from all three branches are fused to produce five segmentation maps of OASS.

\noindent\textbf{SFOASS.}
Considering privacy and storage limitations, this work introduces a more practical task, \textit{i.e.}, SFOASS. 
In the adaptation phase, the source pinhole domain $\mathcal{D}^{pin}$ is locked and inaccessible.
Given a source model $F^{pin}$ well-trained from $\mathcal{D}^{pin}$, and $\mathcal{D}^{pan}$, the goal of SFOASS is to adapt the source model $F^{pin}$ with only unlabeled panoramic data $\{x_{i}^{pan}\}_{i=1}^{N^{pan}}$, obtaining a target model $F^{pan}$ that performs well in the target panoramic domain. (For ease of expression, the superscript ``$pan$'' for target panoramic domain will be omitted henceforth.)

\textit{How can models adapt to panoramic domains without source data, while expanding the field and depth of view?}
To answer this, we propose a novel method called \textbf{UNconstrained Learning Omni-Context Knowledge (UNLOCK)}. It can further enhance the performance of panoramic seamless prediction while improving occlusion-aware reasoning. 
Specifically, as shown in Fig.~\ref{fig_arc}, in response to the challenge that existing self-training methods are ineffective for the SFOASS task, we designed an \textbf{Omni Pseudo-Labeling Learning (OPLL)} approach. This approach leverages all predictions to generate omni soft labels. These generated labels serve as the ``key'' to the proposed UNLOCK.
Furthermore, we propose a brand-new \textbf{Amodal-Driven Contextual Learning (ADCL)} strategy, which effectively resolves the conflict between the real object shapes and contextual knowledge for the occluded object. 
Ultimately, the ADCL holds the ``key'' and performs the ``unlocking" action, completing the source-free adaptation to the panoramic domain.

\subsection{Omni Pseudo-Labeling Learning}
In the SFOASS, only unlabeled target panoramic images $\{x_{i}\}_{i=1}^{N}$ are accessible. Common solutions~\cite{cao2024towards, akkaya2022self, karim2023c} typically involve generating pseudo-labels of target images for training. 
Pseudo-label generation usually entails filtering model predictions based on pre-defined thresholds.
However, we found that directly using the filtered pseudo-labels for self-training in the SFOASS led to performance degradation for instance and amodal instance branches compared to the source model. 
The OASS task consists of a pixel-wise semantic branch and two object-wise instance-level branches (instance and amodal instance) based on region proposals. 
For the instance-level branches, each object is assigned a binary mask for binary cross-entropy loss. However, inaccurate pseudo-labels can hinder learning. When object parts are mistakenly labeled as background, even if the model correctly detects the objects, the pseudo-labels may force it to incorrectly classify them as absent. 

\begin{figure}[htp]
\vskip -2ex
    \begin{subfigure}{1\linewidth}
    \centering
    \includegraphics[width=0.9\linewidth]{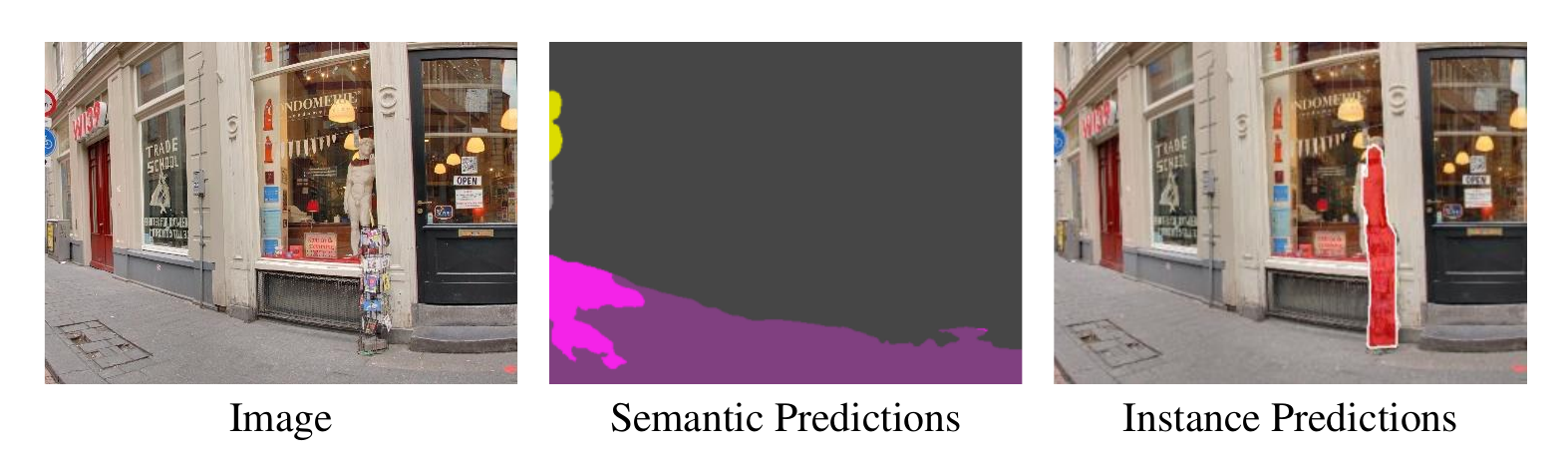}
    \vskip -0.5ex
    \caption{Incorrect predictions (cyclists in \textcolor{red}{red}) of the \textit{Stuff} class.}
    \label{fig_thing}
  \end{subfigure}    
  \begin{subfigure}{1\linewidth}
    \centering
    \includegraphics[width=0.9\linewidth]{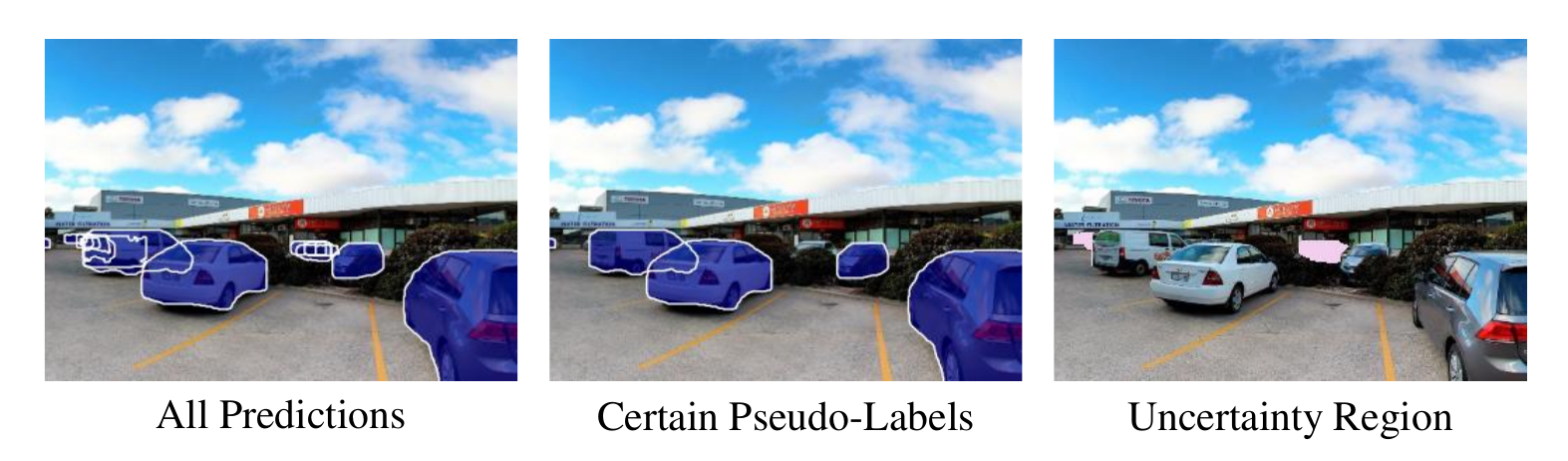}
    \vskip -0.5ex
    \caption{Omni pseudo-label generation with CS thresholds.}
    \label{fig_uncer}
  \end{subfigure}   
    \caption{Solution to the challenge of predictions from the source model in OPLL, illustrated with the amodal instance branch.}
    \label{fig_opll}
\vskip-1.5ex
\end{figure}

To address this, we developed the OPLL method, which introduces generating omni pseudo-labels using Class-wise Self-tuning (CS) thresholds and applying uncertainty-guided instance loss.
Since the two instance-level branches perform the same operations, we will introduce one branch as an example below.
Specifically, we first utilized the semantic branch's predictions to impose additional constraints on the predictions of both instance-level branches.
As shown in Fig.~\ref{fig_thing}, we observed that the instance-level branch, constrained by local information, may misclassify the statue in the shop window as a cyclist of \textit{Thing} class. In contrast, the semantic branch, aided by global information, can usually understand the scenes correctly.
For each panoramic image $x_i$, we obtain the \textit{Thing} mask $m^{thi}$ by the semantic prediction, then use it to revise the object mask of instance-level predictions $p_i^{il}, il\in \{ins,ains\}$:
\begin{equation}
	m^{thi}=\left\{ {\begin{array}{c}
			{1,} \\ 
			{0,} 
		\end{array}\begin{array}{l}
			\text{if}~\text{argmax}p^{sem}\in C^{thi} \\ 
			\text{else}
	\end{array}} \right.
    ,
\end{equation}
\begin{equation}
    \hat{p}_i^{il} = p_i^{il}\cap m^{thi}.
\end{equation}
Then, considering the inconsistent number of objects and the significant imbalance in the source model's capabilities across different \textit{Thing} classes (as shown in \textit{SourceOnly} of Table~\ref{tab_AAP_AP}), we propose a simple and effective CS threshold method for the predictions of the SFOASS task.
For each branch, the CS thresholds $\boldsymbol{\tau}=\{\tau_{c}\}_{c=1}^{C^{thi}}$ is determined by:
\begin{equation}
    \tau_{c}=\text{argmax}_{\tau \in \{{\tau^{fix},\tau^{per}}\}}{N_c^{\tau}},~c\in C^{thi},
    \label{eq2}
\end{equation}
where $N_c^{\tau^{fix}}$ represents the number of objects of class $c$ whose the $score$ exceeds $\tau^{fix}$ across all predictions $\{\hat{p}_i\}_{i=1}^{N}$, and $N_c^{\tau^{per}}$ denotes the number of objects of class $c$ in the top $\tau^{per}$ percentage when the $score$ of all predictions $\{\hat{p}_i\}_{i=1}^{N}$ are sorted in descending order. 
For the semantic branch, the $score$ is replaced with the maximum predicted probability of $p_i^{sem}$. 
This approach generates high-quality pseudo-labels while effectively addressing the issue of low confidence scores for certain classes, caused by the source model $F^{pin}$ inconsistent performance across classes.

Then, we filter out high-quality predictions based on CS thresholds $\boldsymbol{\tau}$ as certain pseudo-labels $\hat{y}_i^{cer}$:

\begin{equation}
    \hat{y}_i^{cer}=\{\hat{p}_i^{(j)}|\hat{p}_i^{(j)}>\boldsymbol{\tau}\},
    \label{eq3}
\end{equation}
where $\hat{p}_i^{(j)}$ represent the prediction of $j$th object in predictions $p_i$ for input image $x_i$. 
To fully utilize the predictions, we treat the remaining masks of the other predictions as uncertainty region $\hat{y}_i^{uncer}$ for the current image:
\begin{equation}
    \hat{y}_i^{uncer} = (1- \mathbf{1}_{\big\{{\sum}_j{\hat{y}_i^{cer}}>0\big\}})\cap(\mathbf{1}_
    {\big\{{\sum}_j{(\hat{p}_i^{(j)}|\hat{p}_i^{(j)}<\boldsymbol{\tau})}>0\big\}}),
    \label{eq_6}
\end{equation}
where $\mathbf{1}_{\{\cdot\}}$ is indicator function. The left-hand side of $\cap$ in Eq.~\ref{eq_6} ensures that the object region of $\hat{y}_i^{cer}$ is not misassigned as the uncertain region.
They forms the omni pseudo-labels $\hat{y}_i=\{ \hat{p}_i^{(1)}, \hat{p}_i^{(2)},..., \hat{y}_i^{uncer}\}$ of the training panoramic image $x_i$.
We believe that all predictions from both instance-level branches of the source model can provide effective knowledge.
For the semantic branch, the $j$ is replaced with the pixel spatial location. 
Finally, for each training panoramic image $x_i$, we obtain three types of omni pseudo-labels $\hat{y}_i^{sem},\hat{y}_i^{ins},\hat{y}_i^{ains}$ for training. During the adaptation process of the target model $F^{pan}$, following~\cite{cao2024oass}, the cross-entropy loss is used for the semantic branch. For the instance-level branches, we introduce a novel uncertain-guided Binary Cross Entropy (BCE) loss $\mathcal{L}_{ur}$, specifically tailored for the SFOASS task.
\begin{equation}
    \mathcal{L}_{ur} = {(1-\hat{y}_i^{uncer})\odot\text{BCE}(F^{pan}(x_i), \hat{y}_i^{cer})}.
\end{equation}
\noindent This approach effectively mitigates prediction errors of the target panoramic data arising from the source pinhole model, allowing the target model to focus on high-quality object samples during the adaptation process.

\subsection{Amodal-Driven Contextual Learning}
Through OPLL, we leverage the source model to generate omni pseudo-labels of the unlabeled panoramic data, obtaining the ``key" for adapting the model to the target domain.
It primarily extracts domain-invariant knowledge from the pinhole-tolerant source model.
We further introduce the ADCL strategy to learn intra-domain knowledge while integrating domain-invariant knowledge. It holds the ``key'' to complete the ``unlocking'' action, and achieves the adaptation to the panoramic images in a source-free manner.

\begin{figure}[!t]
    \centering
    \includegraphics[width=0.9\linewidth]{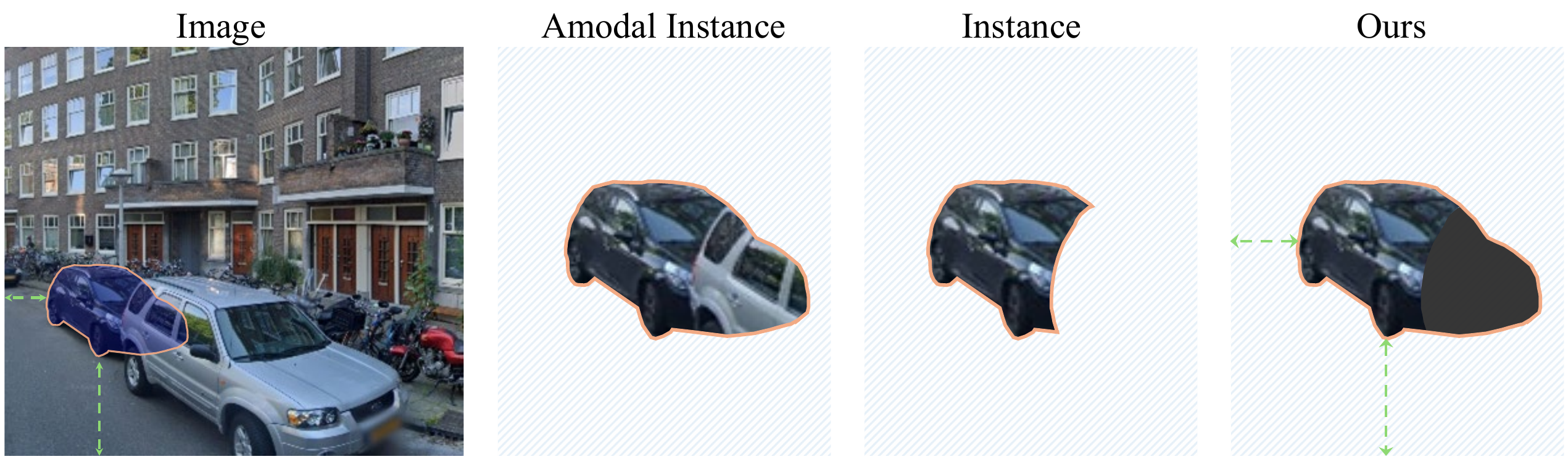}
    \caption{Comparison of object extraction methods: Amodal instance extraction mixes information from other objects, while instance-based extraction may yield incomplete shapes. Our method avoids these issues and preserves spatial awareness.}
    \label{fig_coe}
\end{figure}

Mixing is an effective strategy often used in UDA methods~\cite{akkaya2022self, zhang2018mixup,olsson2021classmix}. The typical approach involves randomly extracting objects from other data and pasting them into the current training image to form new mixed training samples.
However, this approach is not suitable for the OASS model with an amodal instance branch, and it can confuse semantic context and hinder model learning. 
Specifically, since the two instance branches in the OASS model are independent, their predictions are not correlated, leaving us with the option of relying on one of the predictions.
As shown by the car in Fig.~\ref{fig_coe}, the amodal instance prediction captures the complete shape of the car. 
However, this prediction only provides the full mask for the object, meaning that the occluded areas in the mask may inadvertently include parts of other objects.
If these masks are directly used for mixing, incorrect contextual information can be introduced.
On the other hand, relying on instance predictions focuses only on non-occluded regions, making it difficult for the model to learn the full shape of objects. 
This dual challenge complicates the learning process in SFOASS tasks.

\begin{table*}[ht]
\centering
\resizebox{\linewidth}{!}{
\begin{tabular}{c|l|l|l|cccccccccccccccccc}
\hline
 \multicolumn{1}{c}{Task}&\multicolumn{1}{l}{Constraint}&\multicolumn{1}{l}{Method} &\multicolumn{1}{l}{Metric} & \rots{road} & \rots{sidew.} & \rots{build.} & \rots{wall} & \rots{fence} & \rots{pole} & \rots{tr.light} & \rots{tr.sign} & \rots{veget.} & \rots{terrain} & \rots{sky} & \rots{pedes.} & \rots{cyclists} & \rots{car} & \rots{truck} & \rots{ot.veh.} & \rots{van} & \rots{tw.whe.} \\
\hline
\hline
    \multirow{7}{*}{\textbf{APS}} &\multirow{4}{*}{\textcolor[rgb]{0.749, 0.749, 0.749}\faLock~{$(y^{pan})$}}& DATR~\cite{zheng2023look_neighbor} & 20.3  & 51.8  & 09.2  & 59.9  & 11.9  & 12.0  & 02.0  & 00.0  & 03.9  & 64.6  & 14.1  & 70.4  & 11.1  & 00.0  & 39.3  & 00.0  & 03.2  & \textbf{10.1}  & 01.3 \\
    && Trans4PASS~\cite{zhang2022bending} & 22.9  & 53.9  & 14.1  & 69.4  & 19.2  & 11.8  & 03.8  & 00.0  & 05.2  & 67.6  & 16.0  & 77.4  & 15.3  & \textbf{04.2}  & 41.1  & 06.6  & 00.0  & 00.0  & 07.4 \\
    && EDAPS~\cite{saha2023edaps} & 23.1  & 54.9  & 17.0  & 66.9  & 18.8  & 14.5  & 05.8  & 04.0  & 04.6  & 68.2  & 16.0  & 72.8  & 19.0  & 00.0  & 36.7  & \textbf{05.8}  & 04.4  & 00.0  & 07.2 \\
    && UnmaskFormer~\cite{cao2024oass} & \textbf{26.6}  & 61.8  & \textbf{24.7}  & 66.8  & \textbf{20.8}  & \textbf{15.8}  & 05.3  & \textbf{04.3}  & 03.3  & \textbf{69.0}  & \textbf{18.4}  & 79.4  & 20.5  & 03.1  & 44.6  & 12.8  & \textbf{11.3}  & 00.0  & 16.6 \\
    \cdashline{2-22}[2pt/1pt]
    &\textcolor[rgb]{0.5, 0.5, 0.5}\faLock~{$(x^{pan},y^{pan})$} & Source-only & 22.1  & 57.1  & 14.2  & 73.6  & 15.5  & 07.6  & 00.7  & 00.0  & \textbf{10.4}  & 58.3  & 12.4  & 83.1  & 15.2  & 00.0  & 40.3  & 03.8  & 00.0  & 00.0  & 06.1 \\
    \cdashline{2-2}[1.5pt/3pt]
    &\multirow{2}{*}{\textcolor[rgb]{1, 0.8, 0}\faLock~{$(x^{pin},y^{pin},y^{pan})$}}&360SFUDA++~\cite{zheng2024360sfuda++} & 24.5  & 60.3  & 16.8  & 71.7  & 14.1  & 07.7  & 00.0  & 00.0  & 08.4  & 55.6  & 13.6  & 81.3  & 21.5  & 03.2  & \textbf{46.3}  & \textbf{19.4}  & 02.2  & 00.0  & 19.4 \\
    &&\cellcolor[rgb]{0.906, 0.902, 0.902}UNLOCK~(Ours) & \cellcolor[rgb]{0.906, 0.902, 0.902}26.4~($\uparrow$4.3) &\cellcolor[rgb]{0.906, 0.902, 0.902}\textbf{62.7} & \cellcolor[rgb]{0.906, 0.902, 0.902}14.4 & \cellcolor[rgb]{0.906, 0.902, 0.902}\textbf{74.8} & \cellcolor[rgb]{0.906, 0.902, 0.902}20.2 & \cellcolor[rgb]{0.906, 0.902, 0.902}11.2 & \cellcolor[rgb]{0.906, 0.902, 0.902}00.8 & \cellcolor[rgb]{0.906, 0.902, 0.902}00.0 & \cellcolor[rgb]{0.906, 0.902, 0.902}08.0 & \cellcolor[rgb]{0.906, 0.902, 0.902}62.9 & \cellcolor[rgb]{0.906, 0.902, 0.902}18.3 & \cellcolor[rgb]{0.906, 0.902, 0.902}\textbf{84.0} & \cellcolor[rgb]{0.906, 0.902, 0.902}\textbf{21.9} & \cellcolor[rgb]{0.906, 0.902, 0.902}03.6 & \cellcolor[rgb]{0.906, 0.902, 0.902}45.8 & \cellcolor[rgb]{0.906, 0.902, 0.902}\textbf{19.4} & \cellcolor[rgb]{0.906, 0.902, 0.902}00.0 & \cellcolor[rgb]{0.906, 0.902, 0.902}03.1 & \cellcolor[rgb]{0.906, 0.902, 0.902}\textbf{24.4} \\
    \hline
    \multirow{8}{*}{\textbf{PS}} &\multirow{5}{*}{\textcolor[rgb]{0.749, 0.749, 0.749}\faLock~{$(y^{pan})$}}& DATR~\cite{zheng2023look_neighbor} & 19.6  & 50.4  & 09.1  & 59.9  & 11.9  & 12.0  & 02.0  & 00.0  & 03.9  & 64.6  & 14.1  & 70.5  & 12.2  & 00.0  & 38.1  & 00.0  & 03.4  & 00.0  & 01.3 \\
    && Trans4PASS~\cite{zhang2022bending} & 22.7  & 53.9  & 14.1  & 69.4  & 19.2  & 11.8  & 03.8  & 00.0  & 05.2  & 67.6  & 16.0  & 77.4  & 14.6  & \textbf{04.1}  & 38.2  & 06.9  & 00.0  & 00.0  & 07.2 \\
    && UniDAPS~\cite{zhang2022UniDAPS} & 22.7  & \textbf{66.0}  & 09.5  & 66.3  & 17.4  & 14.3  & 04.8  & 00.0  & 06.1  & 67.2  & 16.1  & 72.7  & 08.3  & 00.0  & 27.3  & 14.8  & \textbf{09.2}  & 00.0  & 08.9 \\
    && EDAPS~\cite{saha2023edaps} & 23.1  & 55.0  & 17.1  & 66.8  & 18.7  & 14.5  & \textbf{05.8}  & 04.0  & 04.7  & 68.2  & 16.0  & 72.8  & 19.6  & 00.0  & 37.8  & 01.8  & 04.4  & 00.0  & 07.9 \\
    && UnmaskFormer~\cite{cao2024oass} & \textbf{26.2}  & 61.7  & \textbf{24.7} & 66.8  & \textbf{20.8} & \textbf{15.8} & 05.2  & \textbf{04.3} & 03.3  & \textbf{69.0} & 18.4  & 79.4  & \textbf{20.9} & 03.5  & \textbf{43.0} & 11.3  & 07.6  & 00.0  & 16.0 \\
    \cdashline{2-22}[2pt/1pt]
    &\textcolor[rgb]{0.5, 0.5, 0.5}\faLock~{$(x^{pan},y^{pan})$} & Source-only & 22.3  & 57.8  & 14.2  & 73.8  & 15.5  & 07.6  & 00.7  & 00.0  & 10.4  & 58.3  & 12.4  & 83.2  & 14.9  & 00.0  & 39.1  & 06.0  & 00.0  & 00.0  & 07.7 \\
    \cdashline{2-2}[1.5pt/3pt]
    &\multirow{2}{*}{\textcolor[rgb]{1, 0.8, 0}\faLock~{$(x^{pin},y^{pin},y^{pan})$}} &360SFUDA++~\cite{zheng2024360sfuda++} & 23.7  & 61.3  & 16.8  & 72.2  & 14.1  & 07.6  & 00.0  & 00.0  & \textbf{08.4}  & 55.9  & 13.6  & 81.3  & 18.4  & 00.0  & 40.6  & 16.4  & 02.4  & 00.0  & 17.6 \\
    &&\cellcolor[rgb]{0.906, 0.902, 0.902}UNLOCK~(Ours) &\cellcolor[rgb]{0.906, 0.902, 0.902}25.1~($\uparrow$2.8) &\cellcolor[rgb]{0.906, 0.902, 0.902}60.4 &\cellcolor[rgb]{0.906, 0.902, 0.902}14.4  &\cellcolor[rgb]{0.906, 0.902, 0.902}\textbf{75.7}  &\cellcolor[rgb]{0.906, 0.902, 0.902}20.2  &\cellcolor[rgb]{0.906, 0.902, 0.902}11.2  &\cellcolor[rgb]{0.906, 0.902, 0.902}00.7  &\cellcolor[rgb]{0.906, 0.902, 0.902}00.0  &\cellcolor[rgb]{0.906, 0.902, 0.902}08.0  &\cellcolor[rgb]{0.906, 0.902, 0.902}63.6  &\cellcolor[rgb]{0.906, 0.902, 0.902}18.3  &\cellcolor[rgb]{0.906, 0.902, 0.902}\textbf{84.0}  &\cellcolor[rgb]{0.906, 0.902, 0.902}16.9  &\cellcolor[rgb]{0.906, 0.902, 0.902}00.0  &\cellcolor[rgb]{0.906, 0.902, 0.902}40.0  &\cellcolor[rgb]{0.906, 0.902, 0.902}\textbf{18.5}  &\cellcolor[rgb]{0.906, 0.902, 0.902}00.0  &\cellcolor[rgb]{0.906, 0.902, 0.902}00.0  &\cellcolor[rgb]{0.906, 0.902, 0.902}\textbf{19.4} \\
    \hline
    \multirow{8}{*}{\textbf{SS}} &\multirow{5}{*}{\textcolor[rgb]{0.749, 0.749, 0.749}\faLock~{$(y^{pan})$}}&  DATR~\cite{zheng2023look_neighbor} & 34.9  & 71.9  & 27.2  & 70.6  & 22.8  & 36.0  & 23.9  & 00.0  & 04.5  & 77.1  & 37.1  & 80.1  & 51.2  & 02.2  & 70.2  & 08.9  & 06.0  & 11.0  & 27.8 \\
    && Trans4PASS~\cite{zhang2022bending} & 40.7  & 72.8  & 33.8  & 78.4  & 33.5  & 37.1  & 26.6  & 05.3  & 05.4  & 77.4  & 37.9  & 84.7  & \textbf{57.5}  & 04.5  & 76.6  & 19.0  & 17.0  & 12.7  & 51.7 \\
    && UniDAPS~\cite{zhang2022UniDAPS} & 38.5  & 72.3  & 29.2  & 75.8  & 33.9  & 38.9  & 25.9  & 11.4  & 07.5  & 77.6  & 37.4  & 81.4  & 47.6  & 03.0  & 75.6  & 21.3  & 03.4  & 01.3  & 48.9 \\
    && EDAPS~\cite{saha2023edaps} & 40.2  & 74.4  & 35.9  & 77.0  & \textbf{36.5}  & 40.4  & 28.0  & 16.1  & 05.1  & 78.1  & 39.5  & 82.3  & 55.5  & 03.3  & 74.4  & 06.7  & 14.1  & 05.2  & 50.9 \\
    && UnmaskFormer~\cite{cao2024oass} & \textbf{43.7}    & \textbf{76.5} & \textbf{37.8} & 77.1  & 34.7  & \textbf{44.1} & \textbf{28.3} & \textbf{17.8} & 02.8  & \textbf{78.7} & \textbf{41.7} & 85.0  & 57.3  & \textbf{06.0} & 80.6  & 23.5  & \textbf{21.7} & \textbf{18.8} & 53.6 \\
    \cdashline{2-22}[2pt/1pt]
    &\textcolor[rgb]{0.5, 0.5, 0.5}\faLock~{$(x^{pan},y^{pan})$} & Source-only & 38.7  & 73.0  & 29.1  & 82.0  & 31.2  & 31.4  & 18.0  & 00.0  & 16.0  & 74.1  & 33.7  & 88.7  & 50.2  & 04.0  & 80.4  & 18.0  & 11.0  & 04.6  & 50.2 \\
    \cdashline{2-2}[1.5pt/3pt]
    &\multirow{2}{*}{\textcolor[rgb]{1, 0.8, 0}\faLock~{$(x^{pin},y^{pin},y^{pan})$}}&360SFUDA++~\cite{zheng2024360sfuda++} & 39.4  & 74.6  & 32.2  & 81.2  & 30.7  & 31.9  & 18.1  & 00.0  & 15.6  & 72.6  & 38.3  & 89.2  & 50.6  & 04.6  & 80.5  & 20.4  & 11.6  & 05.3  & 51.0 \\
    &&\cellcolor[rgb]{0.906, 0.902, 0.902}UNLOCK~(Ours) &\cellcolor[rgb]{0.906, 0.902, 0.902}41.6~{%
    {($\uparrow$2.9)}}&\cellcolor[rgb]{0.906, 0.902, 0.902}74.7  &\cellcolor[rgb]{0.906, 0.902, 0.902}32.1  &\cellcolor[rgb]{0.906, 0.902, 0.902}\textbf{83.7}  &\cellcolor[rgb]{0.906, 0.902, 0.902}34.7  &\cellcolor[rgb]{0.906, 0.902, 0.902}40.8  &\cellcolor[rgb]{0.906, 0.902, 0.902}18.5  &\cellcolor[rgb]{0.906, 0.902, 0.902}00.0  &\cellcolor[rgb]{0.906, 0.902, 0.902}\textbf{18.9}  &\cellcolor[rgb]{0.906, 0.902, 0.902}76.1  &\cellcolor[rgb]{0.906, 0.902, 0.902}39.9  &\cellcolor[rgb]{0.906, 0.902, 0.902}\textbf{89.5}  &\cellcolor[rgb]{0.906, 0.902, 0.902}53.7  &\cellcolor[rgb]{0.906, 0.902, 0.902}05.8  &\cellcolor[rgb]{0.906, 0.902, 0.902}\textbf{82.2}  &\cellcolor[rgb]{0.906, 0.902, 0.902}\textbf{26.1}  &\cellcolor[rgb]{0.906, 0.902, 0.902}00.8  &\cellcolor[rgb]{0.906, 0.902, 0.902}14.0  &\cellcolor[rgb]{0.906, 0.902, 0.902}\textbf{57.8} \\
    \hline
\end{tabular}
}
\caption{\textbf{Scene Segmentation} results on the \textbf{K2B} benchmark. Metrics are reported as: mAPQ for Amodal Panoptic Segmentation (\textbf{APS}), mPQ for Panoptic Segmentation (\textbf{PS}), and mIoU for Semantic Segmentation (\textbf{SS}). Per-class results are reported as APQ, PQ, and IoU. $\uparrow$ represents the improvement over the baseline of the Source-only method.
$x^{pin}$ and $x^{pan}$ represent the images, and $y^{pin}$ and $y^{pan}$ represent the labels, from the pinhole source and panoramic target domains, respectively.}
\label{table_K2B_SS}
\vskip-1ex
\end{table*}

First, we use OPLL with stricter thresholds ${{\tau}'}^{fix}$ and ${{\tau}'}^{per}$ to build a buffer pool $\mathcal{B}$ of $K$ high-quality object samples, relying solely on predictions from the amodal instance branch to accurately capture object shapes.
\begin{equation}
\begin{aligned}
    \mathcal{B}=&\{o^{(1)},o^{(2)},...,o^{(K)}\},\\
    &\text{where}~o^{(k)}\in\{\hat{y}_i^{ains\rightarrow cer}|({{\tau}'}^{fix}, {{\tau}'}^{per})\}_{i=1}^{N^{pan}}.
\end{aligned}
\end{equation}
The $o^{(k)}$ represents the filtered certain object obtained through OPLL from the panoramic data. Additionally, for each $o^{(k)}=\{o_{ful}^{(k)},o_{ovp}^{(k)}\}$, 
we include not only the mask $o_{ful}$ of the predicted full region but also record the overlapping region $o_{ovp}$ between the object and other objects in the same image, regardless of whether they are occluded. This process results in an amodal-driven object pool $\mathcal{B}$.

Furthermore, we introduce a novel spatial-aware mixing strategy based on the amodal-driven object pool, enriching the diversity of unlabeled panoramic images by generating amodal-driven mixed samples while respecting panoramic layouts. This spatially consistent mixing ensures semantic coherence, aligning distortions in the panoramic layout even after object insertion.
During the adaptation phase, we randomly select $R$ object samples $\mathcal{O}=\{o^{(r)}\}_{r=1}^{R}$ from $\mathcal{B}$ and integrate them into the current training image $x_i$:
\begin{equation}
    \tilde{x}_i = (1-o_{ful}^{(r)}){\odot}x_i + o_{ful}^{(r)}{\odot}x_r,
\end{equation}
where $x_r$ represents the image corresponding to the paste object $o^{(r)}$.
For ambiguous regions of $o^{(r)}$ (\textit{i.e.}, potentially occluded areas), we assign zero pixels during the mixing process while keeping the complete mask as the label according to the $o_{ovp}^{(r)}$. 
\begin{equation}
    \tilde{x}_i =(1-o_{ovp}^{(r)})\odot \tilde{x}_i.
\end{equation}
For the three pseudo-labels ${\tilde{y}_i}^{sem},{\tilde{y}_i}^{ins},{\tilde{y}_i}^{ains}$ of the mixed image $\tilde{x}_i$, we only use $\{o_{ful}^{(r)}\}_{r=1}^{R}$ to modify them accordingly. This strategy preserves the full shape of objects while avoiding confusion about the semantic content of other objects. The zero-pixel ambiguous regions act as a masked strategy to enhance the model's ability to reconstruct occluded areas and infer the complete shape of occluded objects. This improves the model's ability to learn contextual knowledge from the target panoramic domain, improving its understanding of both individual objects and their relationships within the scene.

\section{Experiments}
\label{experiments}

\subsection{Datasets}
In this paper, we extend the Real-to-Real adaptation on the \textit{KITTI360-APS}${\rightarrow}$\textit{BlendPASS} (\textbf{K2B} for short)) benchmark~\cite{cao2024oass} to SFOASS. Moreover, we pioneer Synthetic-to-Real adaptation in both OASS and SFOASS, introducing \textit{AmodalSynthDrive}${\rightarrow}$\textit{BlendPASS} benchmark (\textbf{A2B} for short) to conduct seamless segmentation research from synthetic to real scenarios. 

\noindent\textbf{Source pinhole domain:}
(1) KITTI360-APS~\cite{mohan2022amodal_panoptic_segmentation}:
It extends KITTI360~\cite{Liao2022PAMI} and includes additional annotations for inmodal and amodal instances. 
A total of $12,320$ annotated images ($1408{\times}376$ pixels) are captured using pinhole cameras across $9$ cities. It covers $11$ \textit{Stuff} classes and $7$ \textit{Thing} classes.
(2) AmodalSynthDrive~\cite{sekkat2023amodalsynthdrive}:
The first synthetic dataset applicable to the OASS task, generated using CARLA~\cite{dosovitskiy2017carla}. It provides images and annotations from four surround-view cameras with a pinhole perspective, with $10,500$ training images ($1920{\times}1080$ pixels) for each viewpoint. It cover $11$ \textit{Stuff} classes and $7$ \textit{Thing} classes.

\noindent\textbf{Target panoramic domain:}
The BlendPASS dataset~\cite{cao2024oass} consists of $2,000$ unlabeled training panoramic images and $100$ labeled test panoramic images ($2048{\times}400$ pixels), captured in real driving scenes across $40$ cities on multiple continents. It covers $11$ \textit{Stuff} and $8$ \textit{Thing} classes.

\begin{figure}[h]
    \centering
    \vskip-2ex
    \includegraphics[width=1\linewidth]{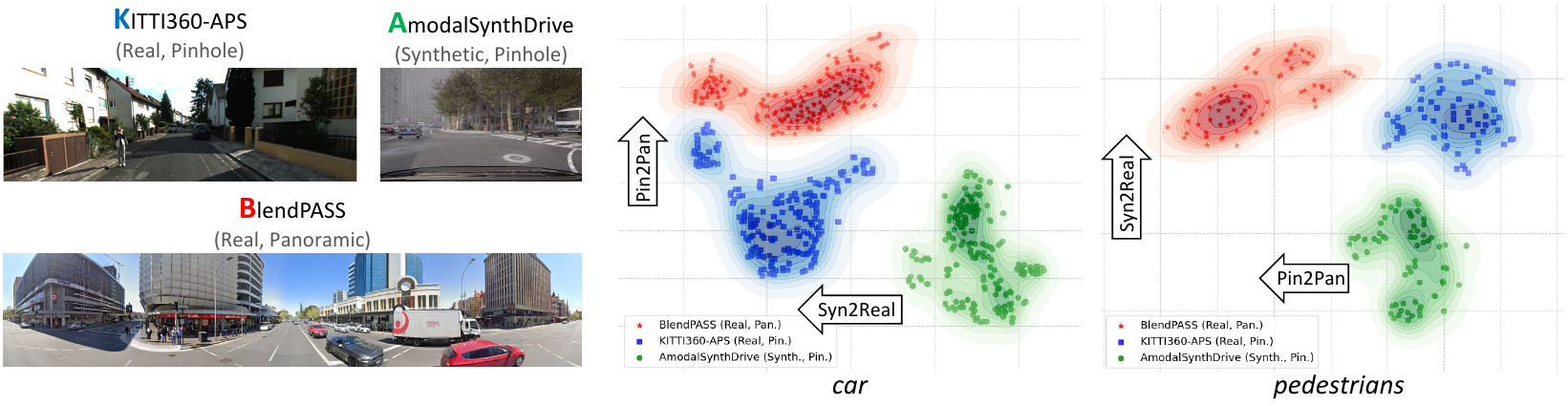}
    
    \caption{Example images from the datasets~\cite{mohan2022amodal_panoptic_segmentation, sekkat2023amodalsynthdrive,cao2024oass} and the class distribution visualized via t-SNE~\cite{tsne}.}
    \vskip-3ex
    \label{fig:domain_gap}
\end{figure}

\noindent \textbf{Domain gap analysis:} Fig.~\ref{fig:domain_gap} shows the feature distributions of \textit{car} and \textit{pedestrians} between the target domain and two source domains, where there is a large domain gap between different datasets, especially for the \textbf{A2B}.

\begin{table*}[ht]
    \centering
    \begin{minipage}[b]{0.48\linewidth}
        \centering
        \resizebox{\textwidth}{!}{ 
        \begin{tabular}{c|c|l|l|ccccccc}
\hline 
\multicolumn{1}{c}{Task} & \multicolumn{1}{l}{}&\multicolumn{1}{l}{Method} & \multicolumn{1}{l}{Metric} & \rots{pedes.} & \rots{cyclists} & \rots{car} & \rots{truck} & \rots{ot.veh.} & \rots{van} & \rots{tw.whe.} \\
\hline\hline
    \multirow{7}[2]{*}{\textbf{AIS}} &\multirow{4}[2]{*}{\textcolor[rgb]{0.749, 0.749, 0.749}\faLock} & DATR~\cite{zheng2023look_neighbor} & 08.7  & 13.1  & 00.0  & 30.6  & 06.9  & 04.7  & \textbf{01.7}  & 03.8 \\
         && Trans4PASS~\cite{zhang2022bending} & 09.9  & 16.0  & \textbf{00.2}  & 31.7  & 08.3  & 06.0  & 00.4  & 06.4 \\
         && EDAPS~\cite{saha2023edaps} & 10.7  & 15.8  & 00.1  & 30.0  & 09.0  & \textbf{12.2}  & 00.4  & 07.4 \\
         && UnmaskFormer~\cite{cao2024oass} & 10.5  & 16.1  & 00.1  & 34.1  & 12.3  & 02.2  & 00.6  & \textbf{08.2} \\
        \cdashline{2-11}[2pt/1pt]
             &\textcolor[rgb]{0.5, 0.5, 0.5}\faLock& Source-only & 10.2  & 15.2  & 00.0  & 33.4  & 12.6  & 03.3  & 00.4  & 06.8 \\
        \cdashline{2-2}[1.5pt/3pt]
          &\multirow{2}[2]{*}{\textcolor[rgb]{1, 0.8, 0}\faLock}  & 360SFUDA++~\cite{zheng2024360sfuda++} & 10.3  & 15.2  & 00.0  & 33.4  & 13.1  & 03.3  & 00.4  & 06.8 \\
        &&\cellcolor[rgb]{0.906, 0.902, 0.902}UNLOCK~(Ours) & \cellcolor[rgb]{0.906, 0.902, 0.902}\textbf{10.9}~($\uparrow$0.7) &\cellcolor[rgb]{0.906, 0.902, 0.902}\textbf{17.1} & \cellcolor[rgb]{0.906, 0.902, 0.902} 00.0 & \cellcolor[rgb]{0.906, 0.902, 0.902}\textbf{34.9} & \cellcolor[rgb]{0.906, 0.902, 0.902}\textbf{14.6} & \cellcolor[rgb]{0.906, 0.902, 0.902}01.7 & \cellcolor[rgb]{0.906, 0.902, 0.902}00.2 & \cellcolor[rgb]{0.906, 0.902, 0.902}07.9 \\
    \hline
    \multirow{8}[2]{*}{\textbf{IS}} &\multirow{5}[2]{*}{\textcolor[rgb]{0.749, 0.749, 0.749}\faLock} & DATR~\cite{zheng2023look_neighbor} & 08.7  & 14.2  & 00.0  & 31.2  & 07.6  & 03.7  & 00.4  & 03.6 \\
         && Trans4PASS~\cite{zhang2022bending} & 10.0  & 16.5  & 00.0  & 32.2  & 10.2  & 05.3  & 00.2  & 05.6 \\
         && UniDAPS~\cite{zhang2022UniDAPS} & 03.4  & 02.3  & 00.0  & 11.3  & 06.2  & 02.8  & 00.0  & 01.5 \\
         && EDAPS~\cite{saha2023edaps} & 10.3  & 16.6  & 00.0  & 30.8  & 06.5  & \textbf{11.4}  & 00.4  & 06.2 \\
         && UnmaskFormer~\cite{cao2024oass} & 11.1  & 17.6  & 00.0  & 35.2  & 14.2  & 02.9  & \textbf{00.8}  & 07.1 \\
          \cdashline{2-11}[2pt/1pt]
          & \textcolor[rgb]{0.5, 0.5, 0.5}\faLock & Source-only & 10.5  & 15.8  & 00.0  & 33.9  & 13.5  & 03.6  & 00.2  & 06.8 \\
          \cdashline{2-2}[1.5pt/3pt]
          &\multirow{2}[2]{*}{\textcolor[rgb]{1, 0.8, 0}\faLock} & 360SFUDA++~\cite{zheng2024360sfuda++} & 10.5  & 15.7  & 00.0  & 33.9  & 13.3  & 03.6  & 00.2  & 06.8 \\
          &&\cellcolor[rgb]{0.906, 0.902, 0.902}UNLOCK~(Ours) & \cellcolor[rgb]{0.906, 0.902, 0.902}\textbf{11.6}~($\uparrow$1.1) &          \cellcolor[rgb]{0.906, 0.902, 0.902}\textbf{17.7} & \cellcolor[rgb]{0.906, 0.902, 0.902}00.0 & \cellcolor[rgb]{0.906, 0.902, 0.902}\textbf{36.1} & \cellcolor[rgb]{0.906, 0.902, 0.902}\textbf{16.3} & \cellcolor[rgb]{0.906, 0.902, 0.902}03.4 & \cellcolor[rgb]{0.906, 0.902, 0.902}00.1 & \cellcolor[rgb]{0.906, 0.902, 0.902}\textbf{07.4} \\
    \hline
\end{tabular}
}
    \end{minipage}
    \hfill
    \begin{minipage}[b]{0.48\linewidth}
        \centering
        \resizebox{\textwidth}{!}{ 
        \begin{tabular}{c|c|l|l|ccccccc }
\hline 
\multicolumn{1}{c}{Task}&\multicolumn{1}{l}{}&\multicolumn{1}{l}{Method} &  \multicolumn{1}{l}{Metric} & \rots{person} & \rots{rider} & \rots{car} & \rots{truck} & \rots{bus} & \rots{motor.} & \rots{bicycle} \\

\hline\hline
    \multirow{7}[2]{*}{\textbf{AIS}} & \multirow{4}[2]{*}{\textcolor[rgb]{0.749, 0.749, 0.749}\faLock} &DATR~\cite{zheng2023look_neighbor} & 08.1  & 14.1  & \textbf{00.4}  & 23.5  & 09.6  & 04.3  & 04.7  & 00.0 \\
          && Trans4PASS~\cite{zhang2022bending} & 09.2  & 13.0  & 00.2  & 24.7  & \textbf{13.0}  & 07.6  & 05.7  & 00.0 \\
          && EDAPS~\cite{saha2023edaps} & 09.5  & 16.7  & 00.2  & 25.2  & 11.6  & 06.6  & 06.1  & 00.0 \\
          && UnmaskFormer~\cite{cao2024oass} & 10.2  & 16.6  & 00.1  & 26.5  & 12.8  & 08.6  & \textbf{06.4}  & 00.0 \\
          \cdashline{2-11}[2pt/1pt]
          &\textcolor[rgb]{0.5, 0.5, 0.5}\faLock & Source-only & 09.7  & 16.0  & 00.0  & 24.9  & 11.9  & \textbf{09.1}  & 05.6  & 00.0 \\
          \cdashline{2-2}[1.5pt/3pt]
          &\multirow{2}[2]{*}{\textcolor[rgb]{1, 0.8, 0}\faLock} & 360SFUDA++~\cite{zheng2024360sfuda++} & 09.6  & 16.2  & 00.0  & 25.3  & 11.7  & 08.9  & 05.2  & 00.0 \\
          &&\cellcolor[rgb]{0.906, 0.902, 0.902}UNLOCK~(Ours) & \cellcolor[rgb]{0.906, 0.902, 0.902}\textbf{10.0}~($\uparrow$0.3) & \cellcolor[rgb]{0.906, 0.902, 0.902}\textbf{17.3} & \cellcolor[rgb]{0.906, 0.902, 0.902}00.0 & \cellcolor[rgb]{0.906, 0.902, 0.902}\textbf{28.9} & \cellcolor[rgb]{0.906, 0.902, 0.902}09.2 & \cellcolor[rgb]{0.906, 0.902, 0.902}08.3 & \cellcolor[rgb]{0.906, 0.902, 0.902}\textbf{06.4} & \cellcolor[rgb]{0.906, 0.902, 0.902}00.0 \\
    \hline
    \multirow{8}[2]{*}{\textbf{IS}} & \multirow{5}[2]{*}{\textcolor[rgb]{0.749, 0.749, 0.749}\faLock}& DATR~\cite{zheng2023look_neighbor} & 08.0  & 14.9  & \textbf{00.4} & 25.3  & 09.2  & 01.0  & 04.9  & 00.0 \\
          && Trans4PASS~\cite{zhang2022bending} & 09.0  & 15.0  & 00.1  & 27.2  & 11.8  & 04.1  & 05.0  & 00.0 \\
          && UniDAPS \cite{zhang2022UniDAPS}  & 00.3  & 00.8  & 00.0  & 00.8  & 00.1  & 00.0  & 00.5  & 00.0 \\
          && EDAPS~\cite{saha2023edaps} & 09.5  & 16.4  & 00.1  & 28.2  & 10.7  & 05.3  & 05.8  & 00.0 \\
          && UnmaskFormer~\cite{cao2024oass} & 09.6  & 17.0  & 00.1  & 28.4  & \textbf{11.9} & 03.8  & 05.9  & 00.0 \\
          \cdashline{2-11}[2pt/1pt]
          & \textcolor[rgb]{0.5, 0.5, 0.5}\faLock & Source-only & 09.3  & 17.4  & 00.0  & 26.9  & 11.6  & 04.4  & 04.9  & 00.0 \\
          \cdashline{2-2}[1.5pt/3pt]
          & \multirow{2}[2]{*}{\textcolor[rgb]{1, 0.8, 0}\faLock} & 360SFUDA++~\cite{zheng2024360sfuda++} & 09.4  & 17.5  & 00.0  & 27.1  & 11.1  & 05.4  & 04.8  & 00.0 \\
          &&\cellcolor[rgb]{0.906, 0.902, 0.902}UNLOCK~(Ours) &\cellcolor[rgb]{0.906, 0.902, 0.902}\textbf{09.7}~($\uparrow$0.4) &\cellcolor[rgb]{0.906, 0.902, 0.902}\textbf{17.7} & \cellcolor[rgb]{0.906, 0.902, 0.902}00.0 & \cellcolor[rgb]{0.906, 0.902, 0.902}\textbf{29.3} & \cellcolor[rgb]{0.906, 0.902, 0.902}09.0 & \cellcolor[rgb]{0.906, 0.902, 0.902}\textbf{05.9} & \cellcolor[rgb]{0.906, 0.902, 0.902}\textbf{06.1} & \cellcolor[rgb]{0.906, 0.902, 0.902}00.0 \\
    \hline
\end{tabular}
}
    \end{minipage}
    \caption{\textbf{Instance-level Segmentation} results on the \textbf{K2B} (left) and \textbf{A2B} (right) benchmark. Metrics are reported as: mAAP for Amodal Instance Segmentation (\textbf{AIS}) and mAAP for Instance Segmentation (\textbf{IS}). Per-class results are reported as AAP and AP.}
\label{tab_AAP_AP}
\end{table*}

\begin{figure*}[ht!]
    \centering
    \vskip-1ex
    \includegraphics[width=0.95\linewidth]{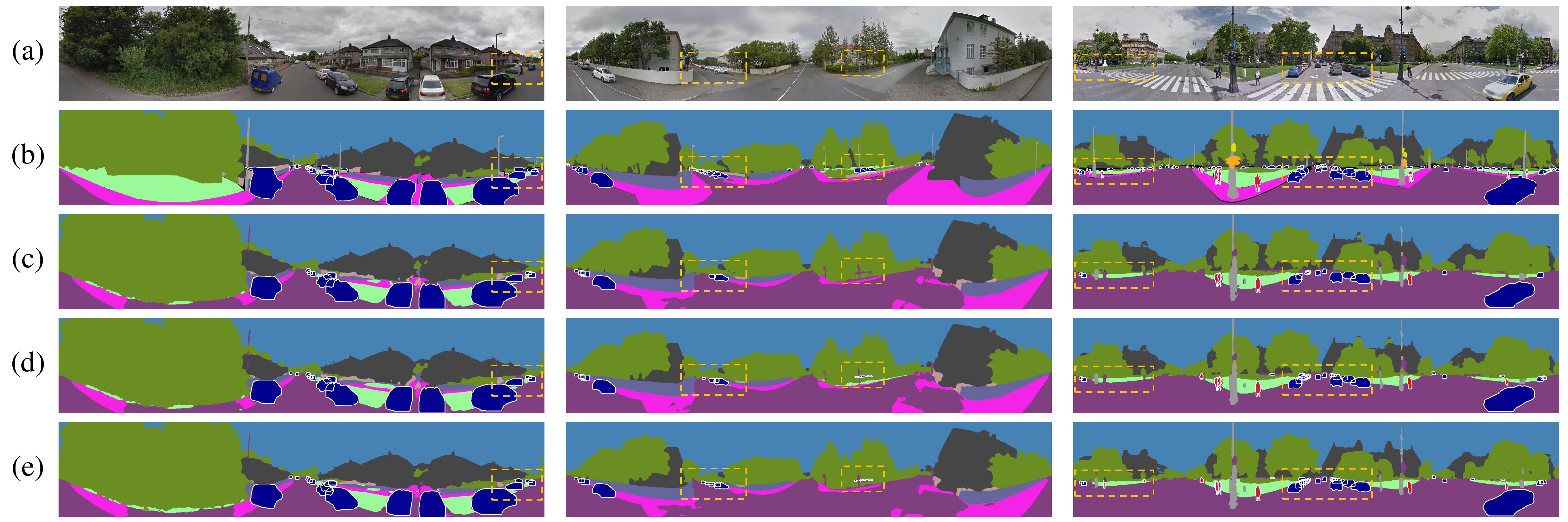}
    \caption{\textbf{Visualization results.} 
    From top to bottom are
    (a) Image, (b) GT, (c) Source-only, and (d) 360SFUDA++~\cite{zheng2024360sfuda++}, and (e) UNLOCK.}
    \label{fig_qualitative}
    \vskip-3ex
\end{figure*}

\subsection{Experiment Setups}
Following~\cite{cao2024oass}, we use the proposed architecture as the backbone of our UNLOCK and the reproduced 360SFUDA++~\cite{zheng2024360sfuda++}. 
In the \textbf{A2B} benchmark, we reproduce existing methods~\cite{zheng2023look_neighbor, zhang2022bending,zhang2022UniDAPS,saha2023edaps,zheng2024360sfuda++} and train the pinhole source model following the framework in~\cite{cao2024oass}.
In the \textbf{K2B} benchmark, we directly use the Source-only model from~\cite{cao2024oass} as the pinhole source model. 
The AdamW optimizer~\cite{loshchilov2017decoupled} is employed, with a learning rate set to $1{\times}10^{-7}$, and a weight decay of $0.01$, followed by polynomial decay. The batch size is set to $4$, with crop sizes of $376{\times}376$ for \textbf{K2B} and $400{\times}400$ for \textbf{A2B}, and both are trained for $10k$ iterations.
In UNLOCK,  $\tau^{fix}$ and $\tau^{per}$ for amodal instance predictions are set to $0.3$ and $0.5$; for instance predictions to $0.5$ and $0.3$; and for semantic predictions, to $0.5$ and $0.8$. For ADCL, they are set to $0.95$ and $0.1$, with the sample object count $R$ set to $10$.
The analysis of these thresholds is provided in the supplementary material.
Our experiments are conducted on a single NVIDIA GTX 3090 GPU using PyTorch.

\subsection{Results of SFOASS}
\noindent\textbf{K2B.}
As shown by the Real-to-Real adaptation results in Table~\ref{table_K2B_SS} and the left side of Table~\ref{tab_AAP_AP}, UNLOCK sets a new benchmark for SFOASS, delivering outstanding performance and effectively addressing the unique challenges of limited FoV, occlusion, and constrained data. 
Compared to the baseline Source-only, UNLOCK outperforms in key metrics for scene understanding including mAPQ, mPQ, and mIoU, with improvements of $+4.3$, $+2.8$, and $+2.9$, respectively.
Even compared to UnmaskFormer~\cite{cao2024oass}, which has access to both source domain and target images, UNLOCK achieves comparable results in mAPQ. 
Notably, in instance-level segmentation, UNLOCK surpasses all existing UDA methods\cite{zheng2023look_neighbor,zhang2022bending,saha2023edaps,cao2024oass}—even without access to source domain data containing $12K$ images-label pairs, reaching the highest mAAP of $10.9$ and mAP of $11.6$, showcasing its robustness and adaptability.
As shown in Fig.~\ref{fig_qualitative}, UNLOCK successfully and precisely segments numerous \textit{Thing} objects with better shape coherence.

\noindent\textbf{A2B.}
The results from another Synthetic-to-Real adaptation scenario, shown in Table~\ref{table_A2B_SS} and the right side of Table~\ref{tab_AAP_AP}, further demonstrate UNLOCK's exceptional performance. 
Compared to 360SFUDA++~\cite{zheng2024360sfuda++}, which is specifically designed for panoramic data, our method significantly outperforms across all SFOASS metrics. 
This success is driven by the OPLL pseudo-label generation strategy, which is tailored specifically for SFOASS. 
The instance-level segmentation results highlight a clear advantage for our method, reflecting its enhanced accuracy in segmenting individual object instances within complex scenes.
\begin{table*}[htbp]
\centering
\resizebox{\linewidth}{!}{
\begin{tabular}{c|l|l|l|cccccccccccccccccc}
\hline 
 \multicolumn{1}{c}{Task}&\multicolumn{1}{l}{Constraint} & \multicolumn{1}{l}{Method} & \multicolumn{1}{l}{Metric} &\rots{road} & \rots{sidew.} & \rots{build.} & \rots{wall} & \rots{fence} & \rots{pole} & \rots{lights} & \rots{sign} & \rots{veget.} & \rots{terrain} & \rots{sky} & \rots{person} & \rots{rider} & \rots{car} & \rots{truck} & \rots{bus} & \rots{motor.} & \rots{bicycle} \\
\hline\hline
    \multirow{7}[2]{*}{\textbf{APS}} &\multirow{4}[2]{*}{\textcolor[rgb]{0.749, 0.749, 0.749}\faLock~{$(y^{pan})$}}& DATR~\cite{zheng2023look_neighbor} & 19.4  & 36.1  & 00.0  & 72.4  & 00.0  & 01.3  & 02.4  & 00.0  & 07.1  & 64.8  & 00.0  & 88.9  & 08.3  & 02.7  & 31.6  & 18.4  & 00.0  & 14.8  & 00.0 \\
         && Trans4PASS~\cite{zhang2022bending} & 21.5  & 38.7  & 00.0  & 71.9  & 00.0  & \textbf{10.0}  & 03.8  & 00.0  & \textbf{07.4}  & 65.3  & 00.0  & 89.2  & 11.1  & \textbf{03.7}  & 35.3  & 25.4  & 07.6  & 15.9  & \textbf{00.9} \\
         && EDAPS~\cite{saha2023edaps} & 21.7  & 45.5  & 00.0  & \textbf{74.9}  & 00.0  & 08.5  & 04.1  & 00.0  & 06.0  & 63.7  & 00.0  & 89.5  & 12.1  & 02.8  & 35.9  & 20.7  & 09.7  & 17.0  & 00.0 \\
         && UnmaskFormer~\cite{cao2024oass} & \textbf{22.9} & 44.1  & 00.6  & 67.5  & 00.0  & 09.6  & \textbf{04.3}  & 00.0  & 05.0  & \textbf{67.0}  & 00.0  & \textbf{90.2}  & 16.6  & 00.0  & 38.7  & \textbf{29.2}  & \textbf{17.1} & \textbf{22.2}  & 00.0 \\
          \cdashline{2-22}[2pt/1pt]
          &\textcolor[rgb]{0.5, 0.5, 0.5}\faLock~{$(x^{pan},y^{pan})$}& Source-only & 18.9  & 47.1  & 01.9  & 68.5  & 00.0  & 00.9  & 00.5  & 00.0  & 01.6  & 43.0  & 00.0  & 84.7  & 09.5  & 00.0  & 36.3  & 21.5  & 07.5  & 16.7  & 00.0 \\
          \cdashline{2-2}[1.5pt/3pt]
          &\multirow{2}[2]{*}{\textcolor[rgb]{1, 0.8, 0}\faLock~{$(x^{pin},y^{pin},y^{pan})$}} & 360SFUDA++~\cite{zheng2024360sfuda++} & 19.4  & 46.6  & 01.9  & 68.1  & 00.0  & 01.0  & 00.5  & 00.0  & 01.7  & 48.2  & 00.0  & 86.1  & \textbf{18.8}  & 00.0  & 38.5  & 18.1  & 06.4  & 14.1  & 00.0 \\
          &&\cellcolor[rgb]{0.906, 0.902, 0.902}UNLOCK~(Ours) &\cellcolor[rgb]{0.906, 0.902, 0.902}21.4~($\uparrow$2.5)  &\cellcolor[rgb]{0.906, 0.902, 0.902}\textbf{50.8}  &\cellcolor[rgb]{0.906, 0.902, 0.902}\textbf{04.7}  &\cellcolor[rgb]{0.906, 0.902, 0.902}73.5  &\cellcolor[rgb]{0.906, 0.902, 0.902}00.0  &\cellcolor[rgb]{0.906, 0.902, 0.902}04.8  &\cellcolor[rgb]{0.906, 0.902, 0.902}00.6  &\cellcolor[rgb]{0.906, 0.902, 0.902}00.0  &\cellcolor[rgb]{0.906, 0.902, 0.902}03.9  &\cellcolor[rgb]{0.906, 0.902, 0.902}49.8  &\cellcolor[rgb]{0.906, 0.902, 0.902}00.0  &\cellcolor[rgb]{0.906, 0.902, 0.902}86.2  &\cellcolor[rgb]{0.906, 0.902, 0.902}17.5  &\cellcolor[rgb]{0.906, 0.902, 0.902}01.2  &\cellcolor[rgb]{0.906, 0.902, 0.902}\textbf{41.4}  &\cellcolor[rgb]{0.906, 0.902, 0.902}20.2  &\cellcolor[rgb]{0.906, 0.902, 0.902}09.3  &\cellcolor[rgb]{0.906, 0.902, 0.902}20.9  &\cellcolor[rgb]{0.906, 0.902, 0.902}00.0 \\
    \hline
    \multirow{8}[2]{*}{\textbf{PS}} &\multirow{5}[2]{*}{\textcolor[rgb]{0.749, 0.749, 0.749}\faLock~{$(y^{pan})$}}& DATR~\cite{zheng2023look_neighbor} & 19.5  & 35.7  & 00.0  & 71.5  & 00.0  & 01.3  & 01.8  & 00.0  & 07.1  & 64.8  & 00.0  & 88.9  & 08.3  & \textbf{05.0}  & 32.2  & 15.8  & 02.4  & 15.5  & 00.0 \\
         && Trans4PASS~\cite{zhang2022bending} & 20.9  & 38.1  & 00.0  & 71.9  & 00.0  & \textbf{10.0}  & 03.8  & 00.0  & 07.4  & 65.3  & 00.0  & 89.2  & 09.6  & 00.0  & 36.5  & 25.0  & 03.3  & 15.9  & 00.0 \\
         && UniDAPS \cite{zhang2022UniDAPS}  & 18.2  & 53.1  & 00.0  & 72.4  & \textbf{01.8}  & 07.8  & \textbf{05.5}  & 00.0  & \textbf{07.7}  & 67.8  & 00.0  & 88.7  & 07.0  & 00.0  & 07.7  & 01.6  & 00.0  & 07.0  & 00.0 \\
         && EDAPS~\cite{saha2023edaps} & 21.9  & 46.7  & 00.0  & \textbf{75.1}  & 00.0  & 08.5  & 04.1  & 00.0  & 06.1  & 63.6  & 00.0  & 89.5  & 11.8  & 02.5  & 36.9  & 23.4  & \textbf{08.9}  & 16.5  & 00.0 \\
         && UnmaskFormer~\cite{cao2024oass} & \textbf{20.7}  & 42.9  & 00.6  & 67.0  & 00.0  & 09.6  & 03.6  & 00.0  & 05.0  & 66.9  & 00.0  & \textbf{90.2}  & \textbf{16.6}  & 00.0  & 37.3  & \textbf{25.9}  & 07.5  & 00.0  & 00.0 \\
          \cdashline{2-22}[1pt/1pt]
          &\textcolor[rgb]{0.5, 0.5, 0.5}\faLock~{$(x^{pan},y^{pan})$}& Source-only & 19.0  & 48.8  & 01.9  & 68.6  & 00.0  & 00.9  & 00.5  & 00.0  & 01.6  & 42.8  & 00.0  & 84.7  & 12.7  & 00.0  & 36.2  & 20.3  & 06.4  & \textbf{16.8}  & 00.0 \\
          \cdashline{2-2}[1.5pt/3pt]
          &\multirow{2}[2]{*}{\textcolor[rgb]{1, 0.8, 0}\faLock~{$(x^{pin},y^{pin},y^{pan})$}} & 360SFUDA++~\cite{zheng2024360sfuda++} & 19.3  & 46.5  & 01.9  & 69.9  & 00.0  & 01.0  & 00.5  & 00.0  & 01.7  & 47.9  & 00.0  & 86.0  & 15.6  & 00.0  & 36.5  & 18.9  & 06.5  & 15.2  & 00.0 \\
         &&\cellcolor[rgb]{0.906, 0.902, 0.902}UNLOCK~(Ours) &\cellcolor[rgb]{0.906, 0.902, 0.902}\textbf{20.4}~($\uparrow$1.4) &\cellcolor[rgb]{0.906, 0.902, 0.902}50.5  &\cellcolor[rgb]{0.906, 0.902, 0.902}\textbf{04.2}  &\cellcolor[rgb]{0.906, 0.902, 0.902}74.0  &\cellcolor[rgb]{0.906, 0.902, 0.902}00.0  &\cellcolor[rgb]{0.906, 0.902, 0.902}04.8  &\cellcolor[rgb]{0.906, 0.902, 0.902}00.6  &\cellcolor[rgb]{0.906, 0.902, 0.902}00.0  &\cellcolor[rgb]{0.906, 0.902, 0.902}03.9  &\cellcolor[rgb]{0.906, 0.902, 0.902}49.2  &\cellcolor[rgb]{0.906, 0.902, 0.902}00.0  &\cellcolor[rgb]{0.906, 0.902, 0.902}86.3  &\cellcolor[rgb]{0.906, 0.902, 0.902}14.3  &\cellcolor[rgb]{0.906, 0.902, 0.902}00.0  &\cellcolor[rgb]{0.906, 0.902, 0.902}\textbf{38.1}  &\cellcolor[rgb]{0.906, 0.902, 0.902}18.8  &\cellcolor[rgb]{0.906, 0.902, 0.902}05.7  &\cellcolor[rgb]{0.906, 0.902, 0.902}16.8  &\cellcolor[rgb]{0.906, 0.902, 0.902}00.0 \\
    \hline
    \multirow{8}[2]{*}{\textbf{SS}} & \multirow{5}[2]{*}{\textcolor[rgb]{0.749, 0.749, 0.749}\faLock~{$(y^{pan})$}}& DATR~\cite{zheng2023look_neighbor} & 31.5  & 59.5  & 06.3  & 77.7  & 00.2  & 09.2  & 22.9  & 02.4  & 10.3  & 76.6  & 00.0  & 93.4  & 37.7  & 01.9  & 68.8  & 20.8  & 01.5  & 54.1  & 24.2 \\
            && Trans4PASS~\cite{zhang2022bending} & 34.3  & 58.9  & 07.6  & 77.3  & 00.2  & 28.7  & 26.7  & 05.2  & \textbf{11.1}  & 76.1  & 00.0  & 93.6  & 35.8  & 01.8  & \textbf{73.6}  & 28.3  & 06.8  & 55.2  & 30.1 \\
            && UniDAPS~\cite{zhang2022UniDAPS}  & 36.0  & 61.5  & 07.5  & 76.6  & 00.8  & 30.4  & \textbf{27.7}  & 10.2  & \textbf{11.1}  & \textbf{77.4}  & 00.0  & 93.6  & \textbf{47.3}  & 02.1  & 49.4  & 25.9  & \textbf{33.0}  & \textbf{62.3}  & \textbf{30.6} \\
            && EDAPS~\cite{saha2023edaps} & \textbf{36.2}  & 62.2  & 10.8  & \textbf{81.1}  & 00.5  & 28.7  & 22.9  & 10.6  & 10.5  & 75.5  & \textbf{00.1}  & 93.3  & 42.7  & \textbf{02.3}  & 72.1  & 33.6  & 18.8  & 58.6  & 26.7 \\
            && UnmaskFormer~\cite{cao2024oass} & 36.1  & 60.6  & 13.3  & 74.2  & 00.9  & \textbf{32.7}  & 21.6  & \textbf{14.2}  & 06.9  & 76.7  & 00.0  & \textbf{93.9}  & 45.2  & 01.2  & 73.3  & \textbf{34.1}  & 22.0  & 56.9  & 22.7 \\
          \cdashline{2-22}[2pt/1pt]
          &\textcolor[rgb]{0.5, 0.5, 0.5}\faLock~{$(x^{pan},y^{pan})$} & Source-only & 31.0  & 67.1  & 21.5  & 76.9  & 03.2  & 17.0  & 17.7  & 02.7  & 06.3  & 64.7  & 00.0  & 90.7  & 40.0  & 01.8  & 69.2  & 19.5  & 04.6  & 47.4  & 08.5 \\
          \cdashline{2-2}[1.5pt/3pt]
          &\multirow{2}[2]{*}{\textcolor[rgb]{1, 0.8, 0}\faLock~{$(x^{pin},y^{pin},y^{pan})$}}& 360SFUDA++~\cite{zheng2024360sfuda++} & 31.2  & 64.8  & 21.3  & 77.7  & \textbf{03.4}  & 16.2  & 18.9  & 02.7  & 06.0  & 66.6  & 00.0  & 91.7  & 41.9  & 02.0  & 67.9  & 19.1  & 04.6  & 49.1  & 07.9 \\
          &&\cellcolor[rgb]{0.906, 0.902, 0.902}UNLOCK~(Ours) &\cellcolor[rgb]{0.906, 0.902, 0.902}34.3~($\uparrow$3.3) &\cellcolor[rgb]{0.906, 0.902, 0.902}\textbf{69.1}  &\cellcolor[rgb]{0.906, 0.902, 0.902}\textbf{25.2}  &\cellcolor[rgb]{0.906, 0.902, 0.902}81.5  &\cellcolor[rgb]{0.906, 0.902, 0.902}00.1  &\cellcolor[rgb]{0.906, 0.902, 0.902}25.1  &\cellcolor[rgb]{0.906, 0.902, 0.902}20.0  &\cellcolor[rgb]{0.906, 0.902, 0.902}03.7  &\cellcolor[rgb]{0.906, 0.902, 0.902}10.5  &\cellcolor[rgb]{0.906, 0.902, 0.902}68.0  &\cellcolor[rgb]{0.906, 0.902, 0.902}00.0  &\cellcolor[rgb]{0.906, 0.902, 0.902}91.6  &\cellcolor[rgb]{0.906, 0.902, 0.902}45.9  &\cellcolor[rgb]{0.906, 0.902, 0.902}01.6  &\cellcolor[rgb]{0.906, 0.902, 0.902}71.4  &\cellcolor[rgb]{0.906, 0.902, 0.902}24.0  &\cellcolor[rgb]{0.906, 0.902, 0.902}07.2  &\cellcolor[rgb]{0.906, 0.902, 0.902}50.8  &\cellcolor[rgb]{0.906, 0.902, 0.902}21.4 \\
    \hline

\end{tabular}
}
\caption{
\textbf{Scene Segmentation} results on the \textbf{A2B} benchmark. Metrics are reported as: mAPQ for Amodal Panoptic Segmentation (\textbf{APS}), mPQ for Panoptic Segmentation (\textbf{PS}), and mIoU for Semantic Segmentation (\textbf{SS}). Per-class results are reported as APQ, PQ, and IoU.
}
\label{table_A2B_SS}
\end{table*}

\subsection{Ablation Study}
We conduct a series of ablation studies on the \textbf{K2B} scenario to evaluate the effectiveness of our method.
\begin{table*}[ht!]
    \centering
    \begin{minipage}[t]{0.32\textwidth}
        \centering
        \resizebox{\linewidth}{!}{
\renewcommand{\arraystretch}{1.}
\begin{tabular}{l|ccccc}
\hline 
& mAPQ & mPQ & mIoU & mAAP& mAP    \\
\hline
Source-only&22.13&22.30&38.65&10.22&\textbf{10.54}\\
Pseudo-Labels&21.90 & 22.07 & \textbf{39.12} & 10.09 & 10.37 \\
All Predictions&22.38 & 22.31 & 38.90 & 10.15 & 10.50\\
\cellcolor[rgb]{0.906, 0.902, 0.902}OPLL (Ours)&\cellcolor[rgb]{0.906, 0.902, 0.902}\textbf{24.71} &\cellcolor[rgb]{0.906, 0.902, 0.902}\textbf{24.00} &\cellcolor[rgb]{0.906, 0.902, 0.902}39.03 &\cellcolor[rgb]{0.906, 0.902, 0.902}\textbf{10.52} &\cellcolor[rgb]{0.906, 0.902, 0.902}10.52\\
\hline
\end{tabular}
}
\caption{Ablation analysis for \textbf{OPLL}.}
\label{tab_ablation_opll}
    \end{minipage}
    \hfill
    \begin{minipage}[t]{0.33\textwidth}
       \centering
\renewcommand{\arraystretch}{1.}
 \resizebox{\linewidth}{!}{
\begin{tabular}{l|ccccc}
\hline
Backbone&mAPQ&mPQ&mIoU&mAAP&mAP\\
\hline
Source-only&22.1&22.3&38.7&10.2&10.5\\
{Trans4PASS~\cite{zhang2022bending}} &26.1&\textbf{25.3}&40.0&\textbf{10.9}&11.3\\
{EDAPS~\cite{saha2023edaps}} &26.3&25.1&40.5&10.7&11.4\\
\cellcolor[rgb]{0.906, 0.902, 0.902}{UnmaskFormer~\cite{cao2024oass}}&\cellcolor[rgb]{0.906, 0.902, 0.902}\textbf{26.4}&\cellcolor[rgb]{0.906, 0.902, 0.902}25.1&\cellcolor[rgb]{0.906, 0.902, 0.902}\textbf{41.6}&\cellcolor[rgb]{0.906, 0.902, 0.902}\textbf{10.9}&\cellcolor[rgb]{0.906, 0.902, 0.902}\textbf{11.6}\\
\hline
\end{tabular}
}
\caption{Ablation analysis for \textbf{backbones}.}
\label{tab_ablation_backbone}
    \end{minipage}
    \hfill
    \begin{minipage}[t]{0.32\textwidth}
        \centering
        \resizebox{\linewidth}{!}{
\renewcommand{\arraystretch}{1.}
\begin{tabular}{cc|ccccc}
\hline 
OPLL&ADCL& mAPQ & mPQ & mIoU & mAAP& mAP    \\
\hline
&&22.13&22.30&38.65&10.22&10.54\\
\checkmark&&24.71 & 24.00 & 39.03 & 10.52 & 10.52\\
&\checkmark&25.84&24.55&40.31&\textbf{10.93}&11.47\\
\rowcolor[rgb]{0.906, 0.902, 0.902}\checkmark&\checkmark&\textbf{26.41}&\textbf{25.07}&\textbf{41.64}&{10.91}&\textbf{11.58}\\
\hline
\end{tabular}
}
\caption{Ablation analysis for \textbf{components}.}
\label{tab_ablation_com}
    \end{minipage}
    \vskip-1.5ex
    \label{tab_ablation_all}
\end{table*}

\begin{table}[ht!]
    \centering
\begin{minipage}[t]{0.63\linewidth}
\resizebox{\linewidth}{!}{
\renewcommand{\arraystretch}{0.9}
\begin{tabular}{c|ccccc}
\hline 
& mAPQ & mPQ & mIoU & mAAP& mAP    \\
\hline
Source-only&22.13&22.30&38.65&10.22&10.54\\
Instance&25.03&24.23&38.93&10.33&10.55\\
Amodal Instance&25.71 & 24.50  & 39.06 & 10.77 & 11.22 \\
AoMix~\cite{cao2024oass} & 25.32 & \textbf{24.67} & 38.04 & 10.62 & 11.26\\
ADCL-Random &24.81&23.90&38.92&10.69&11.19 \\
\rowcolor[rgb]{0.906, 0.902, 0.902}ADCL-Zero (Ours)&\textbf{25.84}&24.55&\textbf{40.31}&\textbf{10.93}&\textbf{11.47}\\
\hline
\end{tabular}
}
\caption{Ablation analysis for \textbf{ADCL}.}
\label{tab_ablation_adcl}
    \end{minipage}
    \hfill
\begin{minipage}[t]{0.35\linewidth}
\centering
\renewcommand{\arraystretch}{0.9}
\resizebox{\linewidth}{!}{
\begin{tabular}{l|c}
\hline
Method&mIoU\\
\hline
SFDA~\cite{liu2021source}&42.70\\
DATC~\cite{yang2022source}&43.06\\
360SFUDA~\cite{zheng2024semantics_style_matter}& 50.12\\
360SFUDA++~\cite{zheng2024360sfuda++}&52.99\\
\cellcolor[rgb]{0.906, 0.902, 0.902}UNLOCK (Ours) &\cellcolor[rgb]{0.906, 0.902, 0.902}\textbf{54.55}\\
\hline
\end{tabular}
}
\caption{\textbf{SS} results on the \textbf{C2D} benchmark.}
\label{tab_ss_c2d}
    \end{minipage}
\vskip-2.5ex
\end{table}

\noindent\textbf{Effectiveness of OPLL.}
An ablation study was conducted to validate the effectiveness of the OPLL. Table~\ref{tab_ablation_opll} presents the comparison of three different approaches for self-training.
Using the final predicted labels from the source model as pseudo-labels or directly using all predictions from the three branches fails to adapt the instance-level branches to the panoramic domain, as evidenced by both mAAP and mAP scores, which are even worse than the no-adaptation Source-only model.
With the OPLL designed by us, the knowledge of the source model can be effectively adapted to the target domain by using omni pseudo-labels. Compared with Source-only, OPLL achieves a significant improvement of $2.58$ in mAPQ.

\noindent\textbf{Effectiveness of ADCL.}
Another ablation study was conducted to validate the effectiveness of the ADCL. Table~\ref{tab_ablation_adcl} presents a comparison of different object mixing strategies. Due to incomplete object shapes, instance-based mixing yields limited improvement in mAPQ and mAAP. Directly using amodal instances causes confusion in contextual information.
Unlike AoMix~\cite{cao2024oass}, which zeros out full amodal object regions of the input image, our method in UNLOCK sets only the overlapping regions to zero (ADCL-Zero), effectively preserving contextual cues. Replacing overlaps with random values (ADCL-Random) yields inferior results, while ADCL-Zero achieves the best mAPQ of $25.84$.

\noindent\textbf{Component Ablation.}
As shown in Table~\ref{tab_ablation_com}, by combining OPLL and ADCL, our UNLOCK method achieves the best overall performance, outperforming the Source-only trained on the source pinhole domain with ${+}4.28$ gains in mAPQ. It shows that our methods can successfully adapt to the target panoramic domain under source-free constraints.

\noindent\textbf{Backbone Ablation.}
As shown in Table~\ref{tab_ablation_backbone}, we evaluated the performance of UNLOCK across different backbones to verify the effectiveness of the UNLOCK strategy itself. The results indicate that UNLOCK maintains consistent performance across diverse backbone architectures, confirming its backbone-agnostic nature as an SFDA framework.

\subsection{Results of Panoramic Semantic Segmentation}
To investigate the generalization capability of UNLOCK,
we extend UNLOCK to the Semantic Segmentation (\textbf{SS}) task and evaluate it on the Cityscapes-to-DensePASS (\textbf{C2D}) benchmark. 
As shown in Table~\ref{tab_ss_c2d}, UNLOCK achieves a significant improvement of ${+}1.56$ in mIoU over 360SFUDA++~\cite{zheng2024360sfuda++}. 
Under the pinhole-to-panoramic \textbf{C2D} adaptation scenario, our UNLOCK achieves the new state-of-the-art result of $54.55$ in mIoU.

\section{Conclusion}
\label{conclusion}
In this paper, we address key constraints in comprehensive scene understanding, to achieve seamless segmentation with 360{\textdegree} viewpoint coverage and occlusion-aware reasoning, while adapting without relying on source data and target labels. To this end, we introduce a new task, Source-Free Occlusion-Aware Seamless Segmentation (SFOASS), and propose its first solution UNLOCK. We further applied two SFOASS benchmarks for evaluation: Real-to-Real and Synthetic-to-Real. 
Extensive experimental results demonstrate the effectiveness of the proposed method.

\section*{Acknowledgment}
\label{sec:Acknowledgment}
This work was supported in part by the National Natural Science Foundation of China (No.~62473139, 62027810), the Major Research Plan of the National Natural Science Foundation of China (No.~92148204), the Top Ten Technical Research Projects of Hunan Province (No.~2024GK1010), the Key Program of Natural Science Foundation of Hunan Province, China (No.~2025JJ30024), the National Key Research and Development Program of China (No.~2024YFB4708900), the Hunan Provincial Research and Development Project (No.~2025QK3019), in part by the Open Research Project of the State Key Laboratory of Industrial Control Technology, China (No. ICT2025B20), and the Science and Technology Project of State Grid Corporation of China (SGCC) Co., LTD (No.~5700-202423229A-1-1-ZN).

{
    \small
    \bibliographystyle{ieeenat_fullname}
    \bibliography{main}
}

\clearpage
\section{SFOASS~{\textit{vs}}~OASS}

\begin{figure}[h]
    \begin{subfigure}{1\linewidth}
    \centering
    \includegraphics[width=\linewidth]{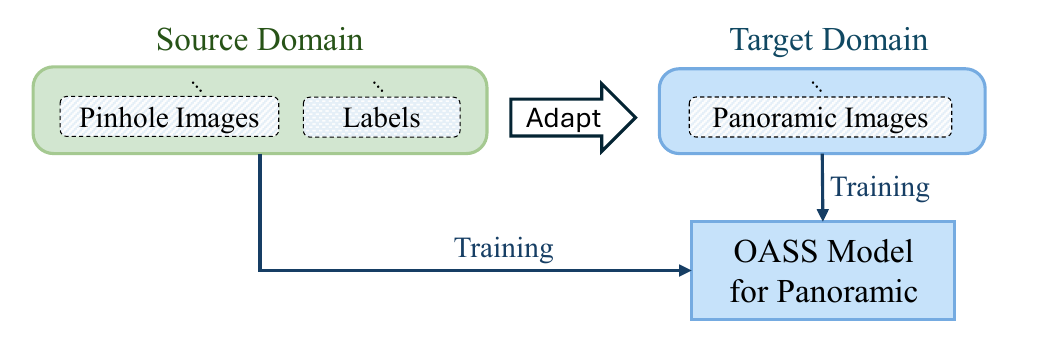}
    \caption{Occlusion-Aware Seamless Segmentation (OASS).}
    \label{sup_fig_oass}
  \end{subfigure}    
  \begin{subfigure}{1\linewidth}
    \centering
    \includegraphics[width=\linewidth]{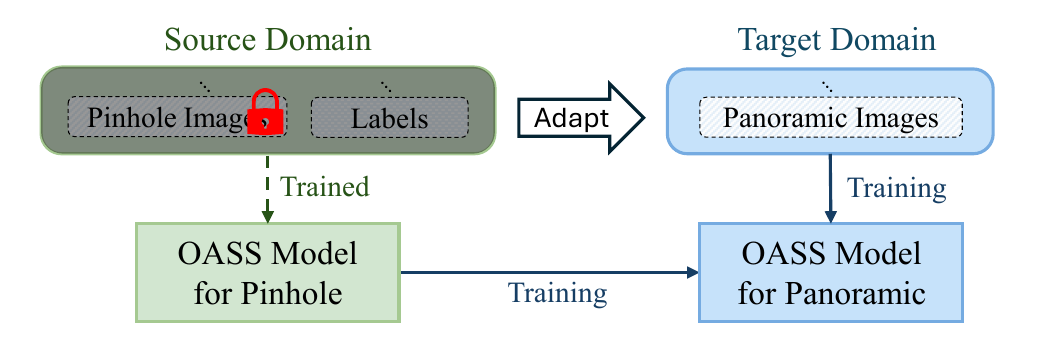}
    \caption{Source-Free Occlusion-Aware Seamless Segmentation (SFOASS).}
    \label{sup_fig_sfoass}
  \end{subfigure}    
    \caption{Comparison of Task Settings.}
\end{figure}

Occlusion-Aware Seamless Segmentation (OASS), introduced by~\cite{cao2024oass}, aims to address the issues of the narrow field of view, occlusion of perspective, and domain gaps in a seamless manner. 
As depicted in Fig.~\ref{sup_fig_oass}, OASS requires the model to transfer from a labeled pinhole domain to the unlabeled panoramic domain, ultimately resulting in an OASS model that performs effectively within the panoramic domain, thereby enabling 360{\textdegree} panoramic perception. In addition, this task introduces amodal-level prediction, which means that the complete shape of the object needs to be segmented regardless of whether the object is occluded or not, in order to solve the problem of perspective occlusion.
The OASS task encompasses five distinct segmentation tasks at once: semantic segmentation, instance segmentation, amodal instance segmentation, panoptic segmentation, and amodal panoptic segmentation.

\begin{figure}
    \centering
    \includegraphics[width=\linewidth]{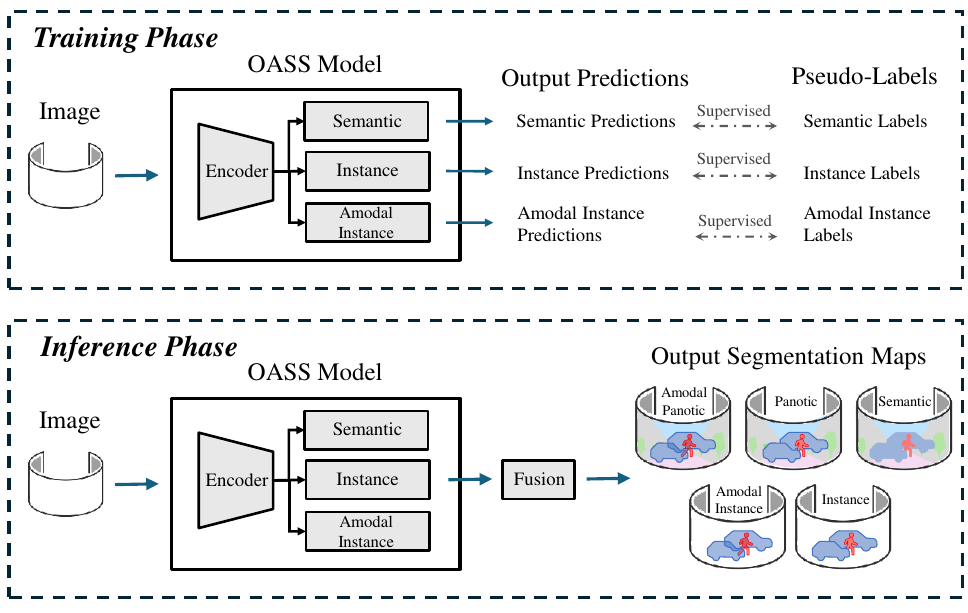}
    \caption{Workflow of the training and inference phases of the OASS model.}
    \label{sup_fig_oassmodel}
    \vskip-3ex
\end{figure}

Typically, the OASS model consists of a shared encoder and three branches: a semantic branch, an instance branch, and an amodal instance branch. 
As illustrated in Fig.~\ref{sup_fig_oassmodel}, during the training phase, the outputs of each branch are supervised using the corresponding labels. 
During inference, in order to obtain all five segmentation maps (amodal panoptic, panoptic, semantic, amodal instance, and instance segmentation) in a single pass, 
the Occlusion-Aware Fusion module~\cite{cao2024oass} combines the pixel-level class predictions from the semantic branch with the instance-level object predictions from the two instance-level branches, generating the final segmentation maps.

However, as a task based on Unsupervised Domain Adaptation (UDA), OASS requires simultaneous access to the data of both the source domain and the target domain during the adaptation process. This presents challenges in scenarios with data privacy and commercial restrictions, where sometimes the data of the source domain is usually prohibited from being accessed.
To address these limitations, we further introduce Source-Free Occlusion-Aware Seamless Segmentation (SFOASS), as depicted in Fig.~\ref{sup_fig_sfoass}. 
This more rigorous task extends the seamless segmentation capabilities of OASS while imposing an additional restriction: the source domain images and labels are inaccessible during domain adaptation. SFOASS relies solely on a pre-trained OASS model from the pinhole domain and unlabeled panoramic images from the target domain. The absence of source domain data during adaptation presents unique challenges, including the lack of explicit domain alignment and the inability to directly address domain-specific biases between the pre-trained source model and unlabeled target images. As such, SFOASS requires the development of innovative strategies to overcome these hurdles and ensure effective adaptation to the target domain. 

To address these challenges, we propose UNconstrained Learning Omni-Context
Knowledge (UNLOCK) framework, a novel framework that introduces Omni Pseudo-Labeling Learning (OPLL) to leverage the knowledge embedded in the pre-trained OASS model. At the same time, it incorporates an Amodal-Driven Contextual Learning mechanism (ADCL) to capture the intrinsic knowledge of the target domain, ensuring effective occlusion-aware segmentation performance in the target panoramic domain.

\section{Benchmarks}
In this work, we applied two benchmarks to evaluate the SFOASS task. These benchmarks are based on three datasets: KITTI360-APS~\cite{mohan2022amodal_panoptic_segmentation}, AmodalSynthDrive~\cite{sekkat2023amodalsynthdrive}, and BlendPASS~\cite{cao2024oass}. In accordance with the SFOASS task formulation, both KITTI360-APS and AmodalSynthDrive serve as source domains, with KITTI360-APS consisting of real-world pinhole data and AmodalSynthDrive consisting of synthetic pinhole data. 
BlendPASS serves as the target domain, composed of real-world panoramic data.
Consequently, two \textbf{Pinhole-to-Panoramic} benchmarks are used: KITTI360-APS$\rightarrow$BlendPASS and AmodalSynthDrive$\rightarrow$BlendPASS. The former constitutes a Real-to-Real adaptation scenario, while the latter forms a Synthetic-to-Real adaptation scenario.

For the KITTI360-APS$\rightarrow$BlendPASS benchmark, the annotated classes between the source and target domains are not fully aligned. 
As detailed in~\cite{cao2024oass}, the annotations of the BlendPASS dataset were pre-processed to match the source domain, KITTI360-APS, resulting in $18$ categories: $11$ \textit{Stuff} classes (\textit{road, sidewalk, building, wall, fence, pole, traffic light, traffic sign, vegetation, terrain, and sky}) and $7$ \textit{Thing} classes (\textit{car, pedestrians, cyclists, two-wheelers, van, truck, and other vehicles}).

For the newly introduced AmodalSynthDrive$\rightarrow$~BlendPASS benchmark, the class annotations are consistent with those of Cityscapes~\cite{cityscapes}, with the exception that the AmodalSynthDrive dataset does not include the \textit{Bus} class, which is present in BlendPASS. To address this discrepancy, we removed the \textit{Bus} class from the annotations of BlendPASS, resulting in $18$ categories: $11$ \textit{Stuff} classes (\textit{road, sidewalk, building, wall, fence, pole, lights, sign, vegetation, terrain, and sky}) and $7$ \textit{Thing} classes (\textit{person, rider, car, truck, bus, motorcycle, and bicycle}). 

\section{More Experiments Details}
In the Amodal-Driven Contextual Learning (ADCL) of UNLOCK, only object samples with an overlapping area of the object region that is less than half of the full area of the object are stored in the amodal-driven object pool. This criterion ensures that excessive contextual information about the objects is not discarded under the spatial-aware mixing strategy.
During the object mixing process, the occlusion order of pasted object samples is determined based on the order in which they are applied. Specifically, the object pasted later is placed in front of all previously pasted objects in the mixed image.
If an object is fully occluded by other pasted objects, it is removed, as this scenario does not require segmentation in the SFOASS task. In this situation, fully occluded objects are outside the scope of the task, as they cannot be detected or even confirmed to exist, making their segmentation unnecessary for the objectives of SFOASS.
For the final mixed omni pseudo-labels, semantic pseudo-labels are directly replaced by the semantic labels of the pasted objects. 
For instance pseudo-labels and amodal instance pseudo-labels, if the objects originally present in the current image (before any pasting) are completely occluded by the pasted objects, the corresponding instance-level pseudo-labels of those objects are removed (since these objects are entirely occluded in the mixed image after pasting).
These operations are crucial to ensuring that the images mixed by the ADCL strategy retain reasonable contextual information. By carefully managing the occlusion and placement of objects, the proposed ADCL approach helps the model learn diverse contextual information, thereby facilitating adaptation to the unlabeled panoramic data of the target domain.
Both OPLL and ADCL operate as data-level preprocessing strategies. OPLL selects informative pseudo-labeled samples based on object-level predictions, while ADCL manipulates object placement and occlusion relationships through spatial-aware mixing. These operations are designed to construct more effective training data and guide the model to adapt better to the unlabeled target domain without modifying the model architecture itself.

For the reproduction of the 360SFUDA++\cite{zheng2024360sfuda++}, we replicated the key innovations and extended the prototype-based approach, initially applied to the semantic segmentation, to the two instance-level branches commonly found in the OASS model, ensuring that both instance branches benefit from this technique. To ensure the robustness and reliability of our reproduction, we strictly followed the original experimental setup. Through multiple rounds of experimentation and fine-tuning, we determined the optimal configuration, achieving results that closely mirror those of the original work.   
For the retraining of existing UDA methods~\cite{zheng2023look_neighbor,zhang2022bending,saha2023edaps,cao2024oass} on the AmodalSynthDrive$\rightarrow$~BlendPASS benchmark, we utilize the training protocols from~\cite{cao2024oass} and adhere to the hyperparameters specified in the respective papers of each method. This approach ensures consistency with the original experimental setups, allowing for a fair comparison across the different methods.

\section{Analysis for Hyper-parameters}
In UNLOCK, OPLL serves as the key, and ADCL performs the unlocking action. We specifically analyze the effect of the hyperparameters in both OPLL and ADCL on the KITTI360-APS$\rightarrow$BlendPASS benchmark. For the SFOASS task, performance is evaluated using five metrics: mAPQ for amodal panoramic segmentation, mPQ for panoramic segmentation, mIoU for semantic segmentation, mAAP for amodal instance segmentation, and mAP for instance segmentation. Since these metrics do not always vary consistently, we prioritize mAPQ, which evaluates amodal panoramic segmentation (including both \textit{Stuff} and amodal-level \textit{Thing} classes), while also taking into account the overall performance across all metrics. Furthermore, although the three branches are designed to be independent, they share the same encoder for feature extraction. As a result, adjustments made to the hyperparameters of a single branch will indirectly affect the other branches, thereby influencing the performance across all metrics.

\begin{table}[ht]
  \centering
  \resizebox{\linewidth}{!}{ 
    \begin{tabular}{cc|ccccc}
    \hline
    $\tau^{fix}$ &$\tau^{per}$ & \multicolumn{1}{c}{mAPQ} & \multicolumn{1}{c}{mPQ} & \multicolumn{1}{c}{mIoU} & \multicolumn{1}{c}{mAAP} & \multicolumn{1}{c}{mAP} \\
    \hline\hline
    0.1&0.9 & 24.37 & 23.78 & 38.84 & 10.41 & 10.43 \\
    0.2&0.7 & 24.45 & 23.76 & 39.10  & 10.45 & 10.43 \\
    \cellcolor[rgb]{0.906, 0.902, 0.902}0.3&\cellcolor[rgb]{0.906, 0.902, 0.902}0.5 & \cellcolor[rgb]{0.906, 0.902, 0.902}\textbf{24.71} & \cellcolor[rgb]{0.906, 0.902, 0.902}\textbf{24.00}    & \cellcolor[rgb]{0.906, 0.902, 0.902}39.03 & \cellcolor[rgb]{0.906, 0.902, 0.902}\textbf{10.52} & \cellcolor[rgb]{0.906, 0.902, 0.902}10.52 \\
    0.4&0.3 & 24.58 & 23.93 & \textbf{39.13} & 10.42 & 10.49 \\
    0.5&0.2 & 24.28 & 23.78 & 38.92 & 10.47 & \textbf{10.53} \\
    \hline
    \end{tabular}%
    }
    \caption{Performance analysis of using different values for $\tau^{fix}$ and $\tau^{per}$ in omni \textbf{amodal instance} pseudo-labels.}
  \label{sup_tab_opll_ai}%
\end{table}%

 \begin{table}[ht]
  \centering
  \resizebox{0.88\linewidth}{!}{ 
  \begin{tabular}{c|ccccc}
    \hline
    $\tau^{fix}$ & \multicolumn{1}{c}{mAPQ} & \multicolumn{1}{c}{mPQ} & \multicolumn{1}{c}{mIoU} & \multicolumn{1}{c}{mAAP} & \multicolumn{1}{c}{mAP} \\
    \hline
        0.1   & 24.52  & 23.88  & 38.95  & \textbf{10.53} & 10.45  \\
    0.2   & 24.34  & 23.74  & 38.77  & 10.31  & 10.51  \\
    \cellcolor[rgb]{0.906, 0.902, 0.902}0.3   & \cellcolor[rgb]{0.906, 0.902, 0.902}\textbf{24.71} & \cellcolor[rgb]{0.906, 0.902, 0.902}\textbf{24.00} & \cellcolor[rgb]{0.906, 0.902, 0.902}39.03  & \cellcolor[rgb]{0.906, 0.902, 0.902}10.52  & \cellcolor[rgb]{0.906, 0.902, 0.902}\textbf{10.52} \\
    0.4   & 24.44  & 23.81  & \textbf{39.14} & 10.39  & 10.35  \\
    0.5   & 24.31  & 23.70  & 38.93  & 10.50  & 10.38  \\
    \hline
    \end{tabular}%
        }
    \caption{Performance analysis of using different values for $\tau^{fix}$ in omni \textbf{amodal instance} pseudo-labels.}
  \label{sup_tab_opll_ai_fix}%
\end{table}%
 \begin{table}[ht]
  \centering
  \resizebox{0.88\linewidth}{!}{ 
    \begin{tabular}{c|ccccc}
    \hline
    $\tau^{per}$ & \multicolumn{1}{c}{mAPQ} & \multicolumn{1}{c}{mPQ} & \multicolumn{1}{c}{mIoU} & \multicolumn{1}{c}{mAAP} & \multicolumn{1}{c}{mAP} \\
    \hline
     0.9   & 24.54  & 23.79  & 38.71  & \textbf{10.52} & 10.36  \\
      0.7   & 24.43  & 23.79  & 38.80  & 10.46  & \textbf{10.60} \\
      \cellcolor[rgb]{0.906, 0.902, 0.902}0.5   & \cellcolor[rgb]{0.906, 0.902, 0.902}\textbf{24.71} & \cellcolor[rgb]{0.906, 0.902, 0.902}\textbf{24.00} & \cellcolor[rgb]{0.906, 0.902, 0.902}\textbf{39.03} & \cellcolor[rgb]{0.906, 0.902, 0.902}\textbf{10.52} & \cellcolor[rgb]{0.906, 0.902, 0.902}10.52  \\
        0.3   & 24.67  & 23.71  & 38.86  & 10.49  & 10.57  \\
        0.2   & 24.56  & 23.90  & 39.02  & 10.47  & 10.35  \\

    \hline
    \end{tabular}%
        }
    \caption{Performance analysis of using different values for  $\tau^{per}$ in omni \textbf{amodal instance} pseudo-labels.}
  \label{sup_tab_opll_ai_per}%
\end{table}%
\begin{table}[htbp]
  \centering
  
  \resizebox{\linewidth}{!}{ 
   \begin{tabular}{cc|ccccc}
    \hline
    $\tau^{fix}$ &$\tau^{per}$ & \multicolumn{1}{c}{mAPQ} & \multicolumn{1}{c}{mPQ} & \multicolumn{1}{c}{mIoU} & \multicolumn{1}{c}{mAAP} & \multicolumn{1}{c}{mAP} \\
    \hline\hline
    0.3&0.6 & 24.37 & 23.66 & 38.81 & 10.32 & 10.32 \\
    0.4&0.4 & 24.53 & 23.93 & 38.95 & 10.45 & 10.51 \\
    \cellcolor[rgb]{0.906, 0.902, 0.902}0.5&\cellcolor[rgb]{0.906, 0.902, 0.902}0.2 & \cellcolor[rgb]{0.906, 0.902, 0.902}\textbf{24.71} & \cellcolor[rgb]{0.906, 0.902, 0.902}\textbf{24.00} & \cellcolor[rgb]{0.906, 0.902, 0.902}39.03 & \cellcolor[rgb]{0.906, 0.902, 0.902}\textbf{10.52} & \cellcolor[rgb]{0.906, 0.902, 0.902}\textbf{10.52} \\
    0.6&0.1 & 24.41 & 23.74 & \textbf{39.12} & 10.42 & 10.39 \\
    0.7&0.05 & 24.20  & 23.61 & 38.80  & 10.45 & 10.37 \\
    \hline
    \end{tabular}%
    }
    \caption{Performance analysis of using different values for $\tau^{fix}$ and $\tau^{per}$ in omni \textbf{instance} pseudo-labels.}
  \label{sup_tab_opll_i}%
\end{table}%

\begin{table}[htbp]
  \centering
  
  \resizebox{0.88\linewidth}{!}{ 
   \begin{tabular}{c|ccccc}
    \hline
    $\tau^{fix}$ & \multicolumn{1}{c}{mAPQ} & \multicolumn{1}{c}{mPQ} & \multicolumn{1}{c}{mIoU} & \multicolumn{1}{c}{mAAP} & \multicolumn{1}{c}{mAP} \\
    \hline\hline
    0.3   & 24.06 & 23.50 & 38.60 & 10.28 & 10.39 \\
    0.4   & 24.20 & 23.52 & 38.65 & 10.22 & 10.43 \\
    \cellcolor[rgb]{0.906, 0.902, 0.902}0.5   & \cellcolor[rgb]{0.906, 0.902, 0.902}\textbf{24.71} & \cellcolor[rgb]{0.906, 0.902, 0.902}\textbf{24.00} & \cellcolor[rgb]{0.906, 0.902, 0.902}39.03 & \cellcolor[rgb]{0.906, 0.902, 0.902}\textbf{10.52} & \cellcolor[rgb]{0.906, 0.902, 0.902}\textbf{10.52} \\
    0.6   & 24.54 & 23.89 & \textbf{39.20} & 10.46 & 10.37 \\
    0.7   & 24.12 & 23.47 & 38.69 & 10.21 & 10.44 \\
    \hline
    \end{tabular}%

    }
    \caption{Performance analysis of using different values for $\tau^{fix}$ in omni \textbf{instance} pseudo-labels.}
  \label{sup_tab_opll_i_fix}%
\end{table}%
\begin{table}[htbp]
  \centering
  
  \resizebox{0.88\linewidth}{!}{ 
   \begin{tabular}{c|ccccc}
    \hline
    $\tau^{per}$ & \multicolumn{1}{c}{mAPQ} & \multicolumn{1}{c}{mPQ} & \multicolumn{1}{c}{mIoU} & \multicolumn{1}{c}{mAAP} & \multicolumn{1}{c}{mAP} \\
    \hline\hline
    0.6   & 24.15 & 23.51 & 38.71 & 10.35 & 10.46 \\
    0.4   & 24.46 & 23.60 & 38.97 & 10.27 & 10.31 \\
    \cellcolor[rgb]{0.906, 0.902, 0.902}0.2   & \cellcolor[rgb]{0.906, 0.902, 0.902}\textbf{24.71} & \cellcolor[rgb]{0.906, 0.902, 0.902}\textbf{24.00} & \cellcolor[rgb]{0.906, 0.902, 0.902}\textbf{39.03} & \cellcolor[rgb]{0.906, 0.902, 0.902}\textbf{10.52} & \cellcolor[rgb]{0.906, 0.902, 0.902}\textbf{10.52} \\
    0.1   & 24.36 & 23.79 & 38.91 & 10.49 & 10.41 \\
    0.05  & 24.27 & 23.59 & 38.87 & 10.25 & 10.43 \\
\hline
    \end{tabular}%
    }
    \caption{Performance analysis of using different values for $\tau^{per}$ in omni \textbf{instance} pseudo-labels.}
  \label{sup_tab_opll_i_per}%
\end{table}%

\begin{table}[htbp]
  \centering

  \resizebox{\linewidth}{!}{ 
    \begin{tabular}{cc|ccccc}
    \hline
    $\tau^{fix}$ &$\tau^{per}$ & \multicolumn{1}{c}{mAPQ} & \multicolumn{1}{c}{mPQ} & \multicolumn{1}{c}{mIoU} & \multicolumn{1}{c}{mAAP} & \multicolumn{1}{c}{mAP} \\
    \hline\hline
    0.3&0.9 & 24.53 & 23.96 & 39.01 & 10.46 & 10.37 \\
    0.4&0.85 & 24.31 & 23.73 & 39.02 & 10.41 & 10.39 \\
    \cellcolor[rgb]{0.906, 0.902, 0.902}0.5&\cellcolor[rgb]{0.906, 0.902, 0.902}0.8 & \cellcolor[rgb]{0.906, 0.902, 0.902}\textbf{24.71} & \cellcolor[rgb]{0.906, 0.902, 0.902}\textbf{24.00} & \cellcolor[rgb]{0.906, 0.902, 0.902}39.03 & \cellcolor[rgb]{0.906, 0.902, 0.902}\textbf{10.52} & \cellcolor[rgb]{0.906, 0.902, 0.902}10.52 \\
    0.6&0.7 & 24.33 & 23.56 & \textbf{39.05} & 10.25 & \textbf{10.58} \\
    0.7&0.6 & 24.34 & 23.52 & 38.95 & 10.28 & 10.47 \\
     \hline
    \end{tabular}%
    }
  \caption{Performance analysis of using different values for $\tau^{fix}$ and $\tau^{per}$ in omni \textbf{semantic} pseudo-labels.}
  \vskip-2ex
  \label{sup_tab_opll_sem}%
\end{table}%

\begin{table}[htbp]
  \centering

  \resizebox{0.88\linewidth}{!}{ 
    \begin{tabular}{c|ccccc}
    \hline
    $\tau^{fix}$ & \multicolumn{1}{c}{mAPQ} & \multicolumn{1}{c}{mPQ} & \multicolumn{1}{c}{mIoU} & \multicolumn{1}{c}{mAAP} & \multicolumn{1}{c}{mAP} \\
    \hline\hline
    0.3   & 24.43 & 23.62 & 39.01 & 10.29 & 10.48 \\
    0.4   & 24.52 & 23.87 & 38.99 & 10.47 & 10.37 \\
    \cellcolor[rgb]{0.906, 0.902, 0.902}0.5   & \cellcolor[rgb]{0.906, 0.902, 0.902}\textbf{24.71} & \cellcolor[rgb]{0.906, 0.902, 0.902}\textbf{24.00} & \cellcolor[rgb]{0.906, 0.902, 0.902}39.03 & \cellcolor[rgb]{0.906, 0.902, 0.902}\textbf{10.52} & \cellcolor[rgb]{0.906, 0.902, 0.902}10.52 \\
    0.6   & 24.46 & 23.81 & \textbf{39.05} & 10.44 & \textbf{10.63} \\
    0.7   & 24.62 & 23.94 & 38.97 & 10.48 & 10.38 \\

     \hline
    \end{tabular}%
    }
  \caption{Performance analysis of using different values for $\tau^{fix}$ in omni \textbf{semantic} pseudo-labels.}
  \vskip-2ex
  \label{sup_tab_opll_sem_fix}%
\end{table}%

\begin{table}[htbp]
  \centering

  \resizebox{0.88\linewidth}{!}{ 
    \begin{tabular}{c|ccccc}
    \hline
    $\tau^{per}$ & \multicolumn{1}{c}{mAPQ} & \multicolumn{1}{c}{mPQ} & \multicolumn{1}{c}{mIoU} & \multicolumn{1}{c}{mAAP} & \multicolumn{1}{c}{mAP} \\
    \hline\hline
    0.9   & 24.30  & 23.67 & 38.88 & 10.37 & 10.38 \\
    0.85  & 24.40  & 23.72 & 38.99 & 10.42 & \textbf{10.59} \\
    \cellcolor[rgb]{0.906, 0.902, 0.902}0.8   & \cellcolor[rgb]{0.906, 0.902, 0.902}\textbf{24.71} & \cellcolor[rgb]{0.906, 0.902, 0.902}\textbf{24.00} & \cellcolor[rgb]{0.906, 0.902, 0.902}\textbf{39.03} & \cellcolor[rgb]{0.906, 0.902, 0.902}\textbf{10.52} & \cellcolor[rgb]{0.906, 0.902, 0.902}10.52 \\
    0.7   & 24.44 & 23.79 & 39.02 & 10.42 & 10.37 \\
    0.6   & 24.29 & 23.7  & 39.01 & 10.41 & 10.37 \\

     \hline
    \end{tabular}%
    }
  \caption{Performance analysis of using different values for $\tau^{per}$ in omni \textbf{semantic} pseudo-labels.}
  \vskip-2ex
  \label{sup_tab_opll_sem_per}%
\end{table}%

\noindent\textbf{OPLL.} 
We systematically investigated the effects of these two parameters $\tau^{fix}$ and $\tau^{per}$ on the omni pseudo-labels from the amodal instance branch, the instance branch, and the semantic branch.
Each experiment focuses solely on OPLL to clearly evaluate its individual contribution.
As shown in Tables~\ref{sup_tab_opll_ai},~\ref{sup_tab_opll_i},~and~\ref{sup_tab_opll_sem}, only the two thresholds associated with the branch under investigation were varied, while the thresholds for the other branches were held constant at their final adopted values, as indicated by the gray background in the tables. Additionally, another set of experiments was conducted to examine the effect of varying individual thresholds, where one threshold was altered at a time while the others remained fixed, as shown in Tables~\ref{sup_tab_opll_ai_fix},~\ref{sup_tab_opll_ai_per},~\ref{sup_tab_opll_i_fix},~\ref{sup_tab_opll_i_per},~\ref{sup_tab_opll_sem_fix},~and~\ref{sup_tab_opll_sem_per}.
Overall, the performance results across the three tables demonstrate that our method exhibits low sensitivity to these hyperparameters. This robustness can be attributed to two aspects: (1) the omni pseudo-labels, which incorporate knowledge from all predictions while excluding low-quality predictions in the optimization process, and (2) the class-wise self-adjusting threshold mechanism, which dynamically maintains a balance between the number and accuracy of the generated labels for each class.
For the amodal instance shown in Table~\ref{sup_tab_opll_ai}, we observed that different values have a relatively minor impact on the mAP and mAAP of the instance-level branch. This is because the amodal instance branch focuses more on the true shape of the object and is less dependent on its appearance features.
As shown in Tables~\ref{sup_tab_opll_ai_fix} and~\ref{sup_tab_opll_ai_per}, the increase in sample numbers due to lower thresholds or higher percentages improves the mAAP or mAP values. However, this increase in samples also slightly affects the accuracy of the \textit{stuff} category, leading to a decrease in other metrics.
For instance pseudo-labels, we observed that excessively low thresholds (as shown in the first row of Table~\ref{sup_tab_opll_i} and~\ref{sup_tab_opll_i_fix}) negatively affect the mAP and mAAP metrics. Lower thresholds lead to the inclusion of incorrect segmentation of the unoccluded regions of objects, which impairs the model's ability to correctly understand the scene. 
For the semantic pseudo-labels, a higher threshold (as shown in the last row of Table~\ref{sup_tab_opll_sem}) can improve the accuracy of semantic labels. However, this also filters out more low-confidence pixels, which reduces the model's ability to capture rich contextual semantic information, ultimately leading to decreased performance.
As shown in Table~\ref{sup_tab_opll_sem_per}, although a higher percentage introduces more pixels, the noise labels it brings also lead to a decrease in performance.

\begin{table}[htbp]
  \centering
  
  \resizebox{\linewidth}{!}{ 
    \begin{tabular}{cc|ccccc}
      \hline
    $\tau'^{fix}$ &$\tau'^{per}$ & \multicolumn{1}{c}{mAPQ} & \multicolumn{1}{c}{mPQ} & \multicolumn{1}{c}{mIoU} & \multicolumn{1}{c}{mAAP} & \multicolumn{1}{c}{mAP} \\
    \hline\hline
    0.99&0.05 & 24.80 & 24.00 & 38.57 & 10.47 & 10.68 \\
    0.95&0.10 & 24.95 & 24.05 & 39.62 & 10.84 & 11.25 \\
    \cellcolor[rgb]{0.906, 0.902, 0.902}0.90&\cellcolor[rgb]{0.906, 0.902, 0.902}0.15 & \cellcolor[rgb]{0.906, 0.902, 0.902}\textbf{25.84} & \cellcolor[rgb]{0.906, 0.902, 0.902}\textbf{24.55} & \cellcolor[rgb]{0.906, 0.902, 0.902}\textbf{40.31} & \cellcolor[rgb]{0.906, 0.902, 0.902}\textbf{10.93} & \cellcolor[rgb]{0.906, 0.902, 0.902}\textbf{11.47} \\
    0.85&0.20 & 25.74 & \textbf{24.55} & 39.27 & 10.81 & 11.20 \\
    0.80&0.30 & 25.39 & 24.39 & 39.23 & 10.62 & 10.91 \\
    \hline
    \end{tabular}%
    }
    \caption{Performance analysis of varying thresholds $\tau'^{fix}$, $\tau'^{per}$ for amodal-driven object pool of ADCL.}
  \label{sup_tab_adcl_ai}%
\end{table}%
\begin{table}[htbp]
  \centering
\resizebox{0.85\linewidth}{!}{ 
    \begin{tabular}{c|ccccc}
    \hline
          $R$& \multicolumn{1}{c}{mAPQ} & \multicolumn{1}{c}{mPQ} & \multicolumn{1}{c}{mIoU} & \multicolumn{1}{c}{mAAP} & \multicolumn{1}{c}{mAP} \\
          \hline
          \hline
    5     & 24.83 & 24.02 & 39.19 & 10.44 & 10.78 \\
    8     & 25.34 & 24.42 & 39.24 & 10.51 & 10.67 \\
    \cellcolor[rgb]{0.906, 0.902, 0.902}10    & \cellcolor[rgb]{0.906, 0.902, 0.902}\textbf{25.84} & \cellcolor[rgb]{0.906, 0.902, 0.902}\textbf{24.55} & \cellcolor[rgb]{0.906, 0.902, 0.902}40.29 & \cellcolor[rgb]{0.906, 0.902, 0.902}\textbf{10.93} & \cellcolor[rgb]{0.906, 0.902, 0.902}\textbf{11.47} \\
    12    & 25.34 & 24.49 & \textbf{40.30} & 10.85 & 11.25 \\
    15    & 25.40  & 24.47 & 39.05 & 10.70  & 10.76 \\
    \hline
    \end{tabular}%
    }
      \caption{Performance analysis of using different numbers $R$ of object samples in ADCL.}
  \label{sup_tab_adcl_r}%
\end{table}%

\noindent\textbf{ADCL.}
We further analyzed the parameters $\tau'^{fix}$ and $\tau'^{per}$, which control the quality of generated objects in the amodal-driven object pool, as well as the parameter $R$, which determines the number of objects pasted in the spatial-aware mixing strategy.
These experiments are conducted independently of OPLL to isolate the effect of ADCL.
As shown in Table~\ref{sup_tab_adcl_ai}, excessively low thresholds lead to a decline in object quality, resulting in the inclusion of background information or incomplete object shapes. Conversely, high thresholds improve the quality of individual objects but reduce the diversity of objects in the pool. For the number $R$ of pasted objects, as shown in Table~\ref{sup_tab_adcl_r}, too few objects limit the variety available for model training, while an overly large number results in overcrowding, which may obscure important contextual information from the \textit{Stuff} class, thereby hindering the model's ability to capture the broader scene context.
\begin{figure*}[ht!]
    \centering
    {
    \newcolumntype{P}[1]{>{\centering\arraybackslash}p{#1}}
    \resizebox{\linewidth}{!}{
    \begin{tabular}{@{}*{20}{P{0.115\columnwidth}}@{}}
    \textit{Stuff:}
    &{\cellcolor[rgb]{0.5,0.25,0.5}}\textcolor{white}{road}
    &{\cellcolor[rgb]{0.957,0.137,0.91}}sidew.
    &{\cellcolor[rgb]{0.275,0.275,0.275}}\textcolor{white}{build.}
    &{\cellcolor[rgb]{0.4,0.4,0.612}}\textcolor{white}{wall}
    &{\cellcolor[rgb]{0.745,0.6,0.6}}fence
    &{\cellcolor[rgb]{0.6,0.6,0.6}}pole
    &{\cellcolor[rgb]{0.98,0.667,0.118}}tr.light
    &{\cellcolor[rgb]{0.863,0.863,0}}tr.sign
    &{\cellcolor[rgb]{0.42,0.557,0.137}}veget.
    &{\cellcolor[rgb]{0.596,0.984,0.596}}terrain
    &{\cellcolor[rgb]{0.275,0.51,0.706}}sky&   
    \textit{~Thing:}
    &{{\cellcolor[rgb]{0.863,0.078,0.235}}\textcolor{white}{pedes.}}
    &{\cellcolor[rgb]{1,0,0}}\textcolor{black}{cyclists}
    &{\cellcolor[rgb]{0,0,0.557}}\textcolor{white}{car}
    &{\cellcolor[rgb]{0,0,0.275}}\textcolor{white}{truck}
    &{{\cellcolor[rgb]{0,0.235,0.392}}\textcolor{white}{ot.veh.}}
    &{\cellcolor[rgb]{0,0.314,0.392}}\textcolor{white}{van}
    &{{\cellcolor[rgb]{0,0,0.902}}\textcolor{white}{tw.whe.}}\\
    \end{tabular}
    }
    }
    \includegraphics[width=0.9\linewidth]{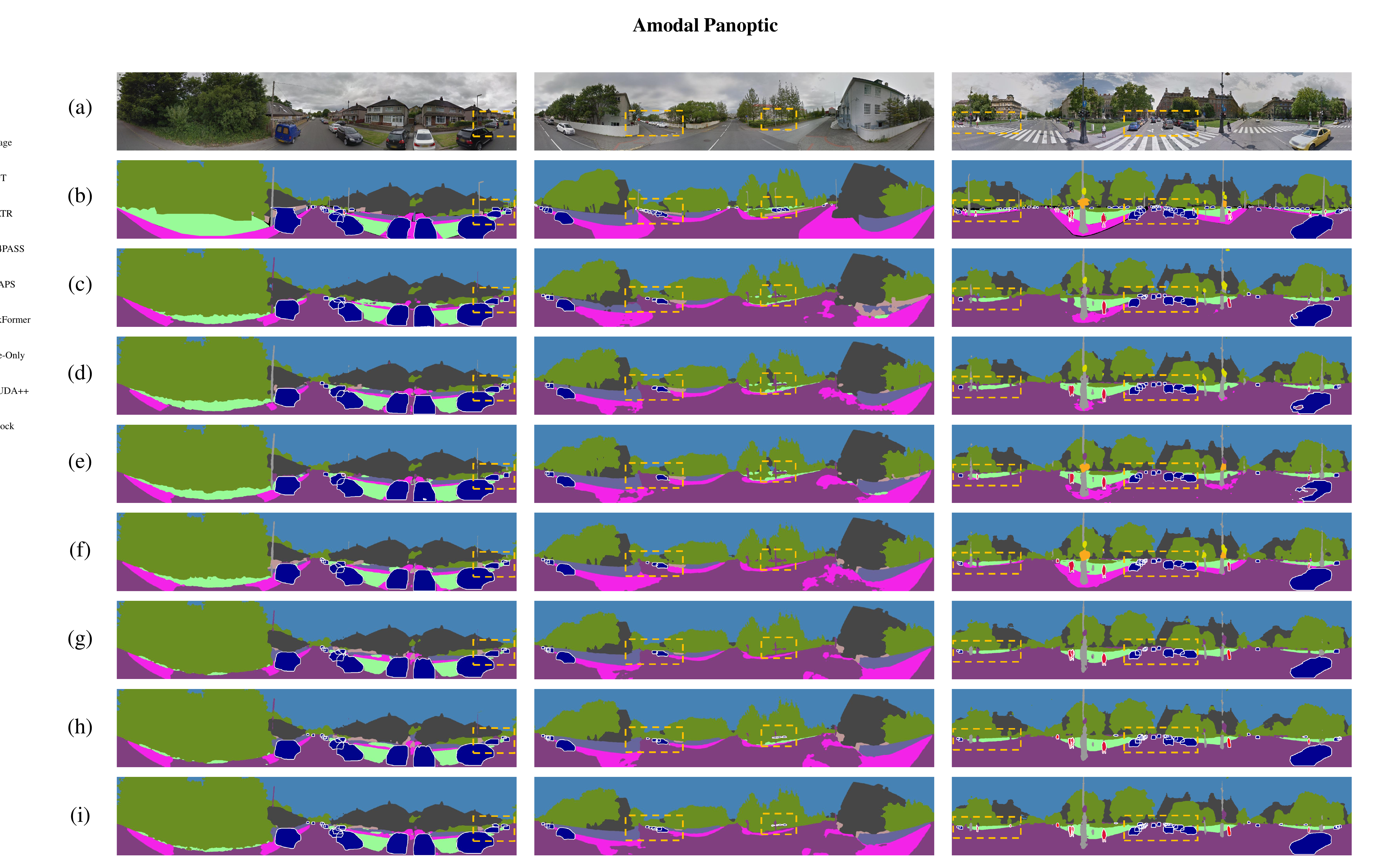}
    \caption{\textbf{Visualization for Amodal Panoptic Segmentation on KITTI360-APS$\rightarrow$BlendPASS benchmark.} 
    From top to bottom are
    (a) Image, (b) GT, (c) DATR~\cite{zheng2023look_neighbor}, (d) Trans4PASS~\cite{zhang2022bending}, (e) EDAPS~\cite{saha2023edaps}, (f) UnmaskFormer~\cite{cao2024oass}, (g) Source-only, (h) 360SFUDA++~\cite{zheng2024360sfuda++}, and (i) UNLOCK (Ours).}
    \label{sup_fig_aps}
    \vskip-2ex
\end{figure*}

\begin{figure*}[h]
    \centering
    {
    \newcolumntype{P}[1]{>{\centering\arraybackslash}p{#1}}
    
    }
    \includegraphics[width=0.9\linewidth]{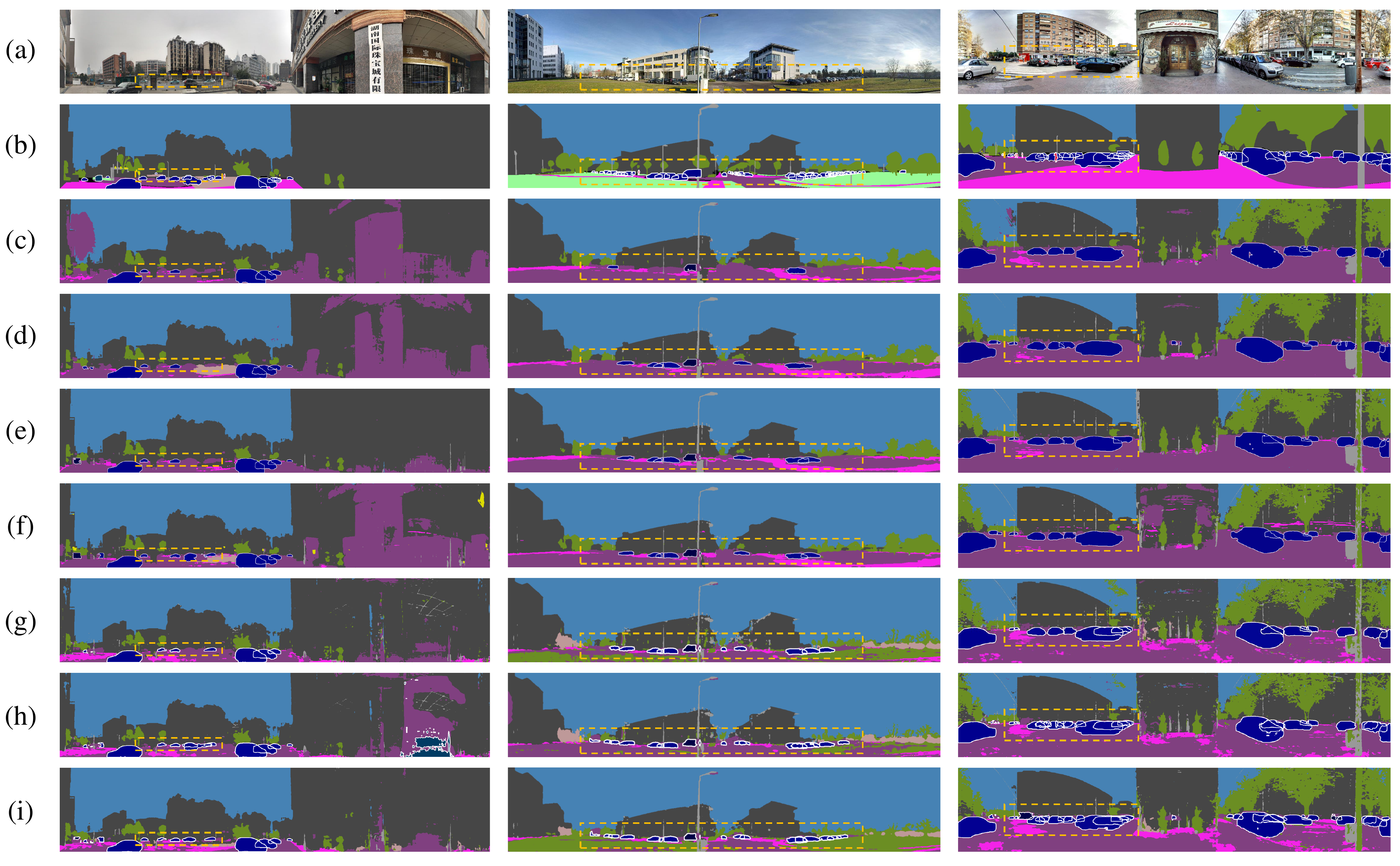}
    \caption{\textbf{Visualization for Amodal Panoptic Segmentation on AmodalSynthDrive$\rightarrow$BlendPASS benchmark.} 
    From top to bottom are
    (a) Image, (b) GT, (c) DATR~\cite{zheng2023look_neighbor}, (d) Trans4PASS~\cite{zhang2022bending}, (e) EDAPS~\cite{saha2023edaps}, (f) UnmaskFormer~\cite{cao2024oass}, (g) Source-only, (h) 360SFUDA++~\cite{zheng2024360sfuda++}, and (i) UNLOCK (Ours).}
    \label{sup_fig_aps_a2b}
\end{figure*}

\section{More Visualization Results}
\subsection{More Visualization Results on SFOASS}
As shown in Fig.~\ref{sup_fig_aps}, we further compare our proposed UNLOCK with existing UDA methods~\cite{zheng2023look_neighbor,zhang2022bending,saha2023edaps,cao2024oass} on KITTI360-APS$\rightarrow$BlendPASS benchmark. The results demonstrate that, even without access to source domain images and labels, UNLOCK achieves comparable, or even surpassing, UDA methods. For example, in the middle column of Fig.~\ref{sup_fig_aps}, UNLOCK detects more cars accurately compared to UnmaskFormer~\cite{cao2024oass}. 
In the right column, UNLOCK excels in identifying and segmenting pedestrians, a result attributed to the designed ADCL method, which facilitates the learning of diverse object samples of the \textit{Thing} class.
We also provided the visualization results for amodal panoptic segmentation on AmodalSynthDrive$\rightarrow$BlendPASS benchmark, as shown in Fig.~\ref{sup_fig_aps_a2b}. 
UNLOCK demonstrates exceptional segmentation performance, particularly when dealing with a high density of \textit{Car} class. Our method identifies more cars with accurate shapes compared to other approaches. This successful adaptation from virtual to real environments not only demonstrates the robustness and generalization of our method but also provides a viable path for applying these advancements in real-world scenarios.

\begin{figure*}[h]
    \centering
    {
    \newcolumntype{P}[1]{>{\centering\arraybackslash}p{#1}}
    }
    \includegraphics[width=\linewidth]{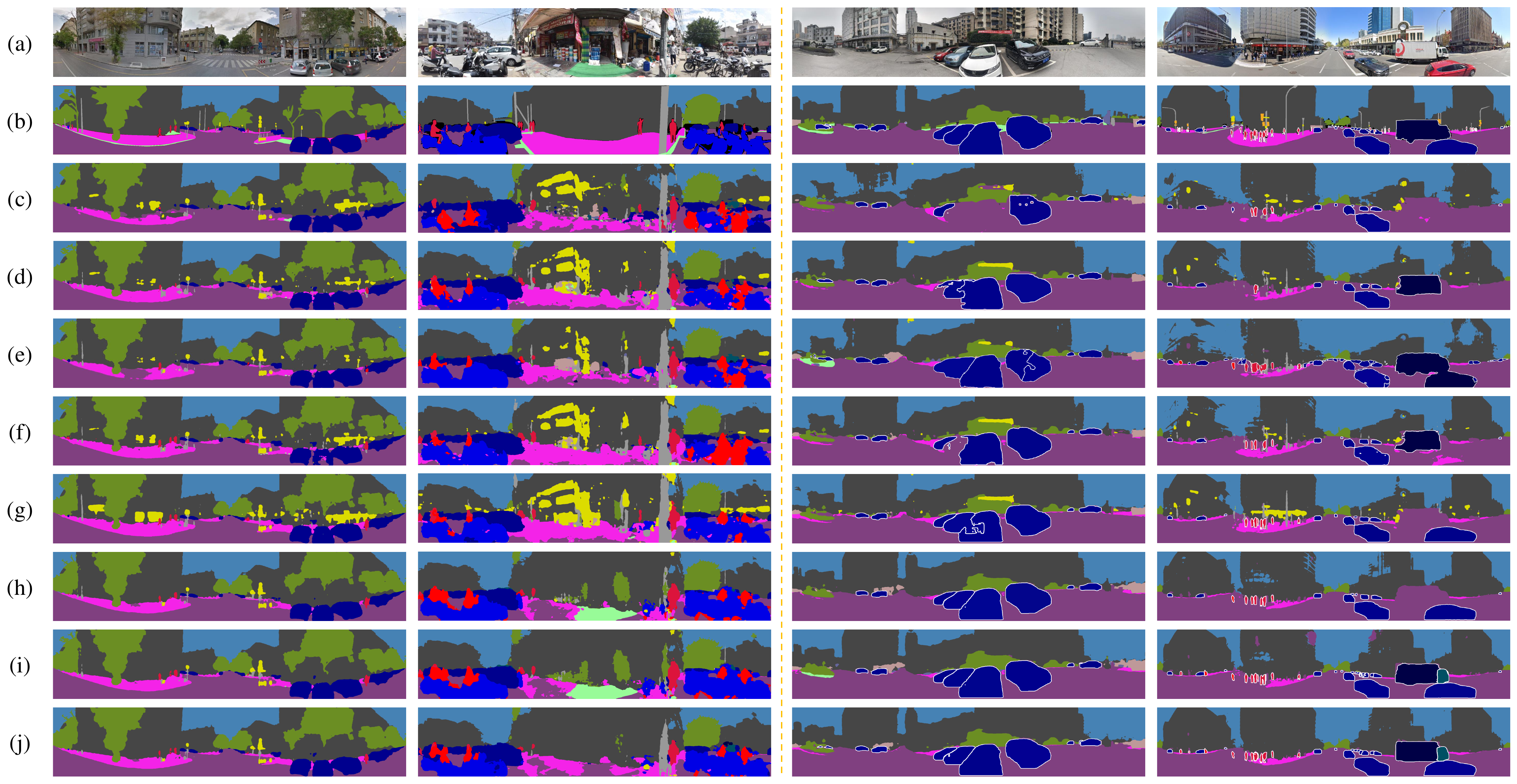}
    \caption{\textbf{Visualization for Semantic (Left) and Panoptic (Right) Segmentation on KITTI360-APS$\rightarrow$BlendPASS benchmark.} 
    From top to bottom are
    (a) Image, (b) GT, (c) DATR~\cite{zheng2023look_neighbor}, (d) Trans4PASS~\cite{zhang2022bending}, (e) UniDAPS~\cite{zhang2022UniDAPS}, (f) EDAPS~\cite{saha2023edaps}, (g) UnmaskFormer~\cite{cao2024oass}, (h) Source-only, (i) 360SFUDA++~\cite{zheng2024360sfuda++}, and (j) UNLOCK (Ours).}
    \label{sup_fig_ss_ps}
\end{figure*}

We also provide the visualization results of semantic segmentation and panoptic segmentation, as shown in Fig.~\ref{sup_fig_ss_ps}. In the left part of Fig.~\ref{sup_fig_ss_ps}, UNLOCK demonstrates superior performance in handling the \textit{Stuff} class. For example, while many UDA methods misidentify elements like billboards on buildings as \textit{Traffic Sign} class, UNLOCK correctly distinguishes these as \textit{Building} class. 
In addition, as shown in the right part of Fig.~\ref{sup_fig_ss_ps}, UNLOCK excels in the panoptic segmentation task, detecting more objects and providing more accurate predictions for each object's visible region compared to other methods (as seen in the third column).

\subsection{Failure Case of UNLOCK}
\begin{figure*}[h]
    \centering
    \includegraphics[width=\linewidth]{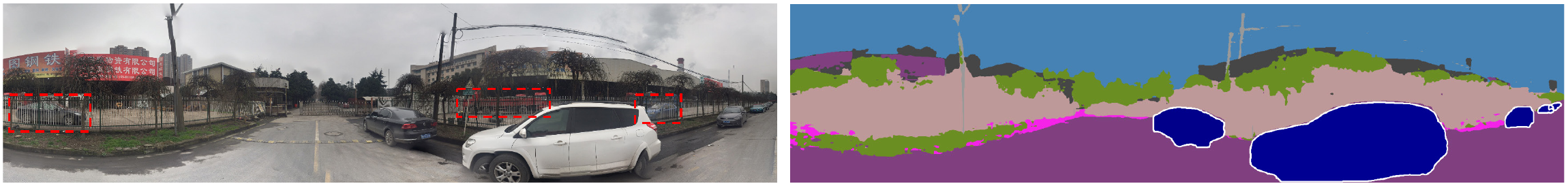}
    \caption{Example of a \textbf{failure case} where UNLOCK misses vehicles partially occluded by sparse fences.} 
    \label{sup_fig_aps_fc}
\end{figure*}

Figure~\ref{sup_fig_aps_fc} illustrates a failure case where vehicles are occluded by sparse fences, leading UNLOCK to miss their presence. Although the vehicles remain partially visible through the gaps, the repetitive vertical patterns of the fence interfere with the visual cues, making it challenging for the model to distinguish the objects from the background. This highlights a limitation of UNLOCK in dealing with structured, non-dense occlusions that disrupt spatial continuity and confuse contextual reasoning.

\subsection{Qualitative analysis of ADCL}
\begin{figure*}[h]
    \centering
    \includegraphics[width=\linewidth]{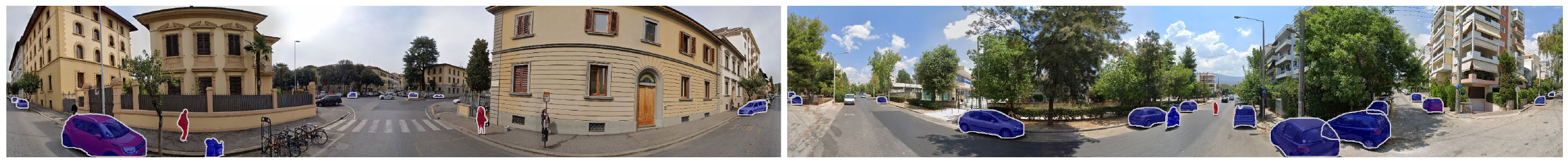}
    \caption{Example of a \textbf{failure case} where UNLOCK misses vehicles partially occluded by sparse fences.} 
    \label{sup_fig_aps_pool}
\end{figure*}

To ensure the reliability of the amodal-driven object pool, we apply stricter adaptive thresholds to filter out high-quality object samples. As shown in Fig.\ref{sup_fig_aps_pool}, we visualize examples from two training images, with the selected high-quality amodal instances highlighted. For instance, the pedestrian in the first column of Fig.\ref{sup_fig_aps_pool} shows accurate segmentation with clear boundaries, while the car in the second column retains a reasonable and complete amodal shape despite being partially occluded by surrounding objects. These results demonstrate that the filtered object pool provides reliable samples, which in turn benefit the spatial-aware mixing strategy in ADCL.

\section{Discussion}
\subsection{Limitations and potential solutions}
The 360{\textdegree} boundaries of panoramic images remain an underexplored aspect in the field of seamless segmentation, posing challenges for accurate perception at extreme viewing angles. Additionally, the interaction between the instance and amodal instance segmentation branches requires further investigation to fully leverage their complementary information.

Future research could explore strategies to mitigate label scarcity in the panoramic domain by incorporating semi-supervised learning techniques into amodal instance segmentation. Moreover, enhancing model robustness and generalizability across diverse panoramic environments could be achieved through domain generalization approaches, enabling improved adaptation to unseen real-world conditions.

In addition, the scalability of the proposed framework with respect to the number of object categories and the complexity of occlusion relationships remains a potential limitation. Since panoramic scenes often contain numerous instances with intricate occlusions, future work should further evaluate and improve the framework’s performance in such large-scale, dense scenarios.

\subsection{Societal Impacts}
On the positive side, SFOASS enhances privacy handling by eliminating the need for access to source-domain images and labels during target-domain adaptation, safeguarding sensitive data. It also facilitates commercial deployment by mitigating data ownership restrictions. Additionally, by relying on a pre-trained model and unlabeled target-domain data, SFOASS helps avoid storage issues related to large training datasets, leading to more efficient resource utilization and supporting sustainable technological development.

However, challenges remain in handling heavily occluded objects and domain gaps, which may lead to misclassifications or biased predictions. In safety-critical applications like autonomous vehicles, such errors could result in accidents. Furthermore, the reliance on pre-trained models without access to source data can limit the adaptability of the system in unfamiliar environments, particularly when the target domain differs significantly from the conditions seen during training.  This could lead to reduced robustness and performance in real-world scenarios, highlighting the need for ongoing validation and refinement.

Therefore, while SFOASS offers promising advancements, its deployment must be carefully managed to mitigate risks and ensure its robustness in real-world scenarios.

\end{document}